\documentclass{optica-article}

\journal{opticajournal} 

\articletype{Research Article}

\usepackage{lineno}
\usepackage{subcaption}
\usepackage{graphicx}
\usepackage{xcolor}
\usepackage{makecell}
\usepackage{booktabs}
\usepackage[normalem]{ulem}
\usepackage{siunitx}
\usepackage{bibunits}
\defaultbibliographystyle{opticajnl} 
\defaultbibliography{references}

\usepackage{longtable}
\usepackage{array}
\usepackage{multirow}

\newcolumntype{L}[1]{>{\raggedright\arraybackslash}p{#1}}
\newcolumntype{C}[1]{>{\centering\arraybackslash}p{#1}}

\newcommand{\rippletableformat}{%
    \small
    \setlength{\tabcolsep}{3pt}%
    \renewcommand{\arraystretch}{1.08}%
    \setlength{\LTleft}{0pt plus 1fill}%
    \setlength{\LTright}{0pt plus 1fill}%
}

\begin{document}
\begin{bibunit}





\title{Motion-guided sparse correction enables expert-quality point tracking across diverse microscopy regimes}

\author{Leonidas Zimianitis,\authormark{1}
Pasindu Thenahandi,\authormark{1,$\dag$}
Kai Buckhalter,\authormark{1,$\dag$}
Dineth Jayakody,\authormark{1}
Julian O. Kimura,\authormark{2,3}
Xinyue Liang,\authormark{2,3}
Karen Cunningham,\authormark{2,3}
Azeem Ahmad,\authormark{4}
Balpreet S. Ahluwalia,\authormark{4,5}
Sampath Jayarathna,\authormark{1}
Nikos Chrisochoides,\authormark{1,6}
Brandon Weissbourd,\authormark{2,3}
and Dushan N. Wadduwage\authormark{1,6,7,*}}

\address{\authormark{1}Department of Computer Science, Old Dominion University, Norfolk, VA 23529, USA\\
\authormark{2}Department of Biology, Massachusetts Institute of Technology, Cambridge, MA 02139, USA\\
\authormark{3}The Picower Institute for Learning and Memory, Massachusetts Institute of Technology, Cambridge, MA 02139, USA\\
\authormark{4}Department of Physics and Technology, UiT--The Arctic University of Norway, Troms\o{} 9037, Norway\\
\authormark{5}Department of Physics, University of Oslo, Oslo 0316, Norway\\
\authormark{6}School of Data Science, Old Dominion University, Norfolk, VA 23529, USA\\
\authormark{7}Department of Physics, Old Dominion University, Norfolk, VA 23529, USA\\
\authormark{$\dag$}These authors contributed equally.}

\email{\authormark{*}dwadduwa@odu.edu}

\begin{abstract*}
Tracking the dynamics of non-canonical biological systems in microscopy videos remains a persistent challenge. Both classical and learning-based trackers depend on expert-reviewed data to be evaluated and adapted, yet exhaustive manual annotation rarely scales to the videos where these tools are needed most. We developed RIPPLE (Refinement Interpolation Platform for Point Location Estimation), which recasts annotation as sparse correction: a user clicks a starting point, RIPPLE proposes a full trajectory, and the user intervenes only where the trajectory drifts. We tested RIPPLE on five challenging microscopy datasets from our laboratories, four from the transparent jellyfish \emph{Clytia hemisphaerica} and one tracking landmarks on rapidly moving sperm. Across these, RIPPLE matched the quality of exhaustive manual annotation while reducing manual clicks by 3 to 25 times across datasets. RIPPLE thereby fills a missing layer between manual annotation and fully automated tracking, enabling immediate quantification of biological dynamics, method benchmarking, and the production of the gold-standard data needed to adapt future automated microscopy trackers.
\end{abstract*}

\section{Introduction}

Modern microscopy now resolves living systems across orders of magnitude in space and time, capturing how molecules, cells, and entire nervous systems change moment to moment \cite{wu2022multiscale,pylvanainen2023live}. Extracting biology from these recordings hinges on a single computational task: following the same object reliably from one frame to the next as the sequence evolves \cite{jaqaman2008robust,chenouard2014objective,sbalzarini2005feature,cheng2022review}. Tracking turns raw video into the quantitative measurements that drive discovery, from receptor and intracellular dynamics at the subcellular scale \cite{jaqaman2008robust,manzo2015review,chenouard2014objective,worth2009live,zhang2021intracellular}, to cell behavior, proliferation, and lineage relationships in time-lapse microscopy \cite{ker2018phase,lugagne2020delta,versari2017long,li2008cell,malin2023automated}, to relating brain-wide activity to behavior in freely moving and deforming nervous systems \cite{atanas2026deep}.

Despite a long history of tracking research, every major class of method depends on the same fragile resource: high-quality, expert-reviewed data \cite{cheng2022review,cao2025rethinking}. Classical and semi-automated packages such as TrackMate \cite{tinevez2017trackmate,ershov2022trackmate}, u-track \cite{jaqaman2008robust,roudot2023u}, and OrganoidTracker \cite{kok2020organoidtracker} require carefully curated reference tracks so methods can be evaluated objectively and tracker outputs compared against trajectories human experts have inspected \cite{chenouard2014objective,ulman2017objective}. Learning-based and retrainable systems, including DeLTA \cite{lugagne2020delta}, Cellpose \cite{stringer2021cellpose}, StarDist \cite{schmidt2018cell}, SLEAP \cite{pereira2022sleap}, and DeepLabCut \cite{nath2019using}, depend on expert-annotated examples for training and fine-tuning \cite{cao2025rethinking}. The newest generation of general-purpose computer-vision trackers extends this dependency further \cite{doersch2023tapir,karaev2024cotracker,cho2024local,harley2022particle,karaev2025cotracker3}. These models can in principle follow any chosen point through large motions, occlusions, and temporary disappearances, making them attractive for dim, crowded, blurry, and deforming microscopy targets that move in and out of focus \cite{spilger2021deep,hansen2018robust,pylvanainen2023live,wu2022multiscale,mavska2023cell}. Yet adapting them to biological video still requires curated specimen-specific datasets that do not exist at the necessary scale \cite{cao2025rethinking,karaev2025cotracker3,zheng2023pointodyssey}. Across paradigms, then, the practical obstacle is not running a tracker; it is generating the data that experts need to evaluate, adapt, and improve the tracker.

Producing such data is expensive. Manual annotation at scale quickly exceeds what individual labs can afford \cite{van2021biological,cao2025rethinking}: even community benchmarks built across multiple institutions still rely on labor-intensive gold-standard labelling \cite{mavska2023cell,mavska2014benchmark}, and focused studies report hundreds of hours of effort despite using semi-automated tools for assistance, with track annotation taking 2.5 hours on average per track \cite{ker2018phase}. To cope, researchers typically chain automated detection or tracking with rounds of manual tuning, curation, and editing \cite{kok2020organoidtracker,sugawara2022tracking,versari2017long,ershov2022trackmate,han2019edetect,van2021biological,wagner2021tracurate}. These hybrid workflows reduce effort relative to fully bespoke annotation, but still require expert review to validate and correct outputs \cite{kok2020organoidtracker,wagner2021tracurate,han2019edetect,padovani2022segmentation}, and they often force users to retune parameters or chain multiple tools whenever the dataset or experimental setting changes \cite{ershov2022trackmate,fukai2023laptrack,fazeli2020automated,padovani2022segmentation}. They also strain on exactly the conditions where they are needed most: low signal-to-noise ratios, motion blur, transient disappearances, out-of-focus motion, deformation, and changing feature appearance routinely break automated trackers \cite{spilger2021deep,hansen2018robust,wu2022multiscale,mavska2023cell}, forcing yet more frequent manual intervention \cite{han2019edetect}. What is missing is a unified framework that lets experts annotate difficult microscopy data efficiently, especially when automated pipelines are unavailable, brittle, or expensive to adapt, while preserving the data quality those pipelines ultimately depend on.

We address this gap with RIPPLE (\underline{R}efinement \underline{I}nterpolation \underline{P}latform for \underline{P}oint \underline{L}ocation \underline{E}stimation), a platform that recasts the central manual task---keeping a point on the right object across time---into a sparse-correction problem. The user clicks a starting point, RIPPLE proposes a full trajectory using image-motion-guided interpolation, and the user only intervenes where the trajectory drifts. Because each correction propagates locally through the surrounding frames, a single click can rapidly correct many frames. We tested RIPPLE on five challenging microscopy datasets we had previously acquired in our laboratories that span the regimes where automated tracking typically struggles. Four are from the transparent jellyfish \emph{Clytia hemisphaerica}, an emerging model for whole-nervous-system imaging in behaving animals \cite{weissbourd2021genetically,chari2021whole,houliston2022past} in which rapid, deformable body motion routinely defeats both classical and deep-learning trackers; the fifth tracks landmarks on rapidly moving sperm. Across all five datasets, RIPPLE matched the quality of exhaustive manual annotation while reducing manual click effort by 3 to 25 times across datasets. TAP-Vid \cite{doersch2022tap} solved the analogous sparse point-tracking problem in general computer vision, and it motivated our sparse-correction workflow. However, TAP-Vid was too slow for our microscopy data: after each correction, it rebuilt trajectories too slowly to support real-time interaction. RIPPLE updates tracks more than 10,000 times faster than TAP-Vid while preserving similar accuracy, making sparse correction responsive enough for expert-guided microscopy annotation. By making expert-quality tracks practical to obtain on non-canonical and difficult microscopy data, RIPPLE simultaneously enables immediate quantification of dynamics, method evaluation, and dataset construction, and provides the gold-standard data needed to adapt the next generation of data-driven tracking models in microscopy.

\begin{figure*}[t]
\centering
\includegraphics[width=\textwidth]{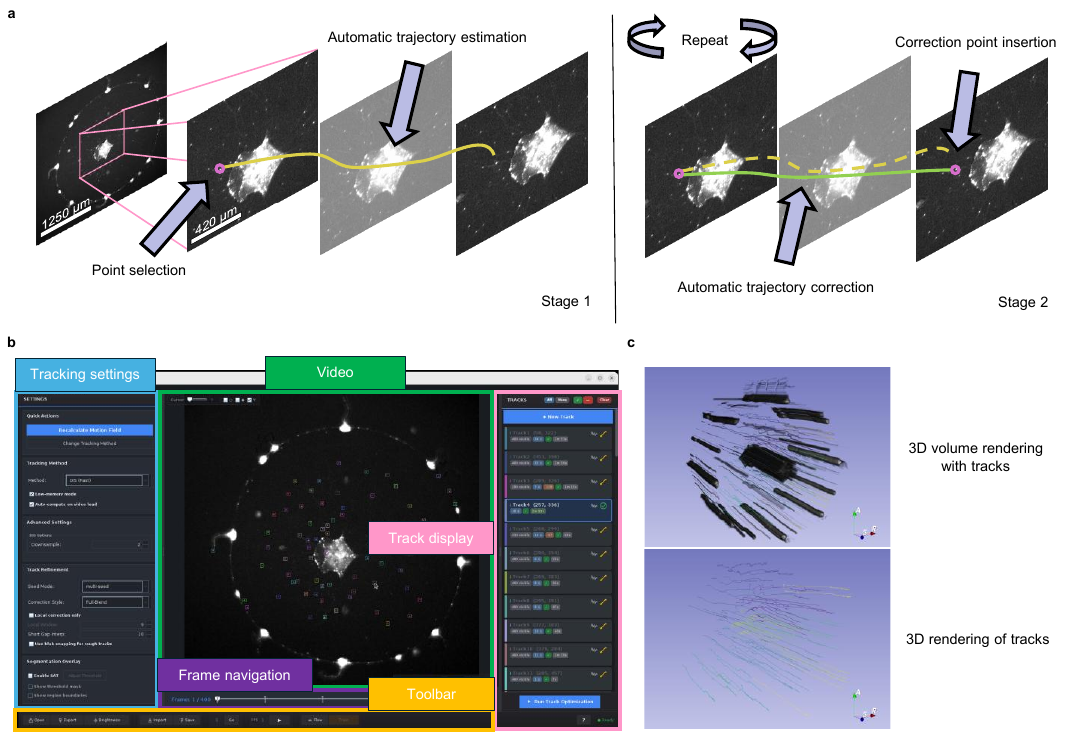}
\caption{\textbf{RIPPLE workflow, interface, and volumetric track visualization.}
\textbf{a}, RIPPLE sparse-correction workflow. A user selects a point, RIPPLE estimates a full trajectory automatically, and later corrective insertions trigger local track updating; this process can be repeated until the trajectory is satisfactory.
\textbf{b}, Graphical interface of RIPPLE, showing tracking settings, video display, frame navigation, toolbar controls, and track list for interactive review and editing.
\textbf{c}, Example 3D visualizations of tracked data, including track overlays on a volume rendering and a rendering of trajectories alone. Together, these panels illustrate RIPPLE as an interactive review-and-correct framework for point tracking rather than a dense frame-by-frame annotation workflow.}
\label{fig:ripple_overview}
\end{figure*}

\section{Results}

\subsection{RIPPLE enables sparse correction through motion-guided interpolation}

RIPPLE makes sparse correction effective by tracking movement forward and backward through time. A user starts a track by placing a point on any frame, and the software propagates that point through the video to produce a trajectory. The user inspects the trajectory and corrects it wherever it drifts; each correction coherently updates the trajectory, and the user repeats until the track is satisfactory. This workflow (Fig.~\ref{fig:ripple_overview}a) hinges on how the software fills in motion between sparse user corrections. To achieve that we used Dense Inverse Search (DIS) pixel motion estimation \cite{kroeger2016fast}. With RIPPLE we introduce a lightweight bidirectional correction update. When the user clicks an initial point, RIPPLE applies the estimated image shift to carry that point forward and backward through the video, frame by frame, generating the full track. The user inspects the track and, if it drifts, clicks the correct position to insert a manual correction. The two nearest manual points then serve as anchors: RIPPLE propagates the earlier anchor forward and the later anchor backward, producing two predicted positions for each intermediate frame. It blends these predictions, weighting each most heavily near its originating anchor and tapering toward the middle. Blending forward and backward flow estimates lets RIPPLE faithfully follow curved, non-rigid, and complex motion paths. Detailed description of the software and the underlying algorithm are presented in the method section.

\subsection{RIPPLE reduces annotation effort across diverse microscopy datasets}

\begin{figure*}[t]
\centering
\includegraphics[width=0.93\textwidth]{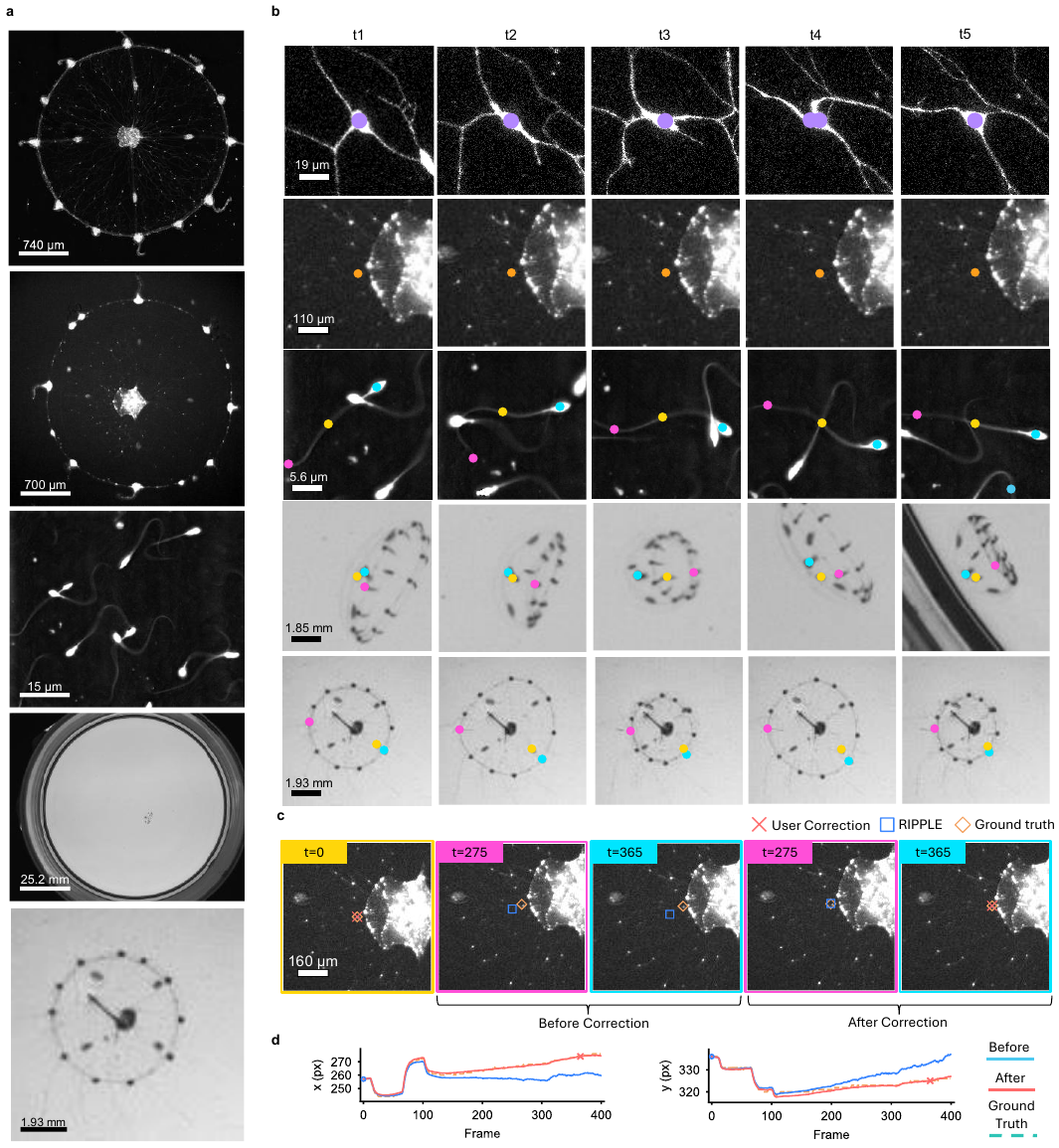}
\caption{\textbf{Dataset diversity and representative sparse-correction behavior.}
\textbf{a}, Representative fields of view from the five microscopy regimes used in this study.
\textbf{b}, Example tracked targets at selected time points, illustrating differences in motion, appearance, deformation, crowding, and ambiguity across datasets.
\textbf{c}, Representative correction event on a neural trajectory. A user correction inserted at a later frame re-aligns the propagated track to the intended target across the intervening interval.
\textbf{d}, Corresponding coordinate traces before and after correction relative to the manual comparison track, showing how sparse intervention can substantially improve temporal agreement without requiring manual annotation of every frame.}
\label{fig:dataset_overview}
\end{figure*}

We evaluated RIPPLE across five microscopy datasets generated in our laboratories, each with different tracked targets and sources of difficulty (Fig.~\ref{fig:dataset_overview}). In the Neural \emph{Clytia} dataset, we tracked individual neurons during rapid body contractions, where rapid body and tentacle deformation temporarily blurred features and neurons inside the mouth obscured one another. In the Pinned \emph{Clytia} dataset, we tracked tentacle bulbs and gonads while the restrained animal underwent strong three-dimensional deformation. In the Freely swimming \emph{Clytia} dataset, we tracked the mouth, gonads, and tentacle bulbs as the animal translated, rotated, and self-occluded, adding global motion and occlusion to local deformation. In the Homeostatic \emph{Clytia} dataset, we tracked individual neurons across sparsely sampled timelapse images, where large temporal gaps made it difficult to identify the same neuron across frames. In the Sperm QPM dataset, we tracked three landmarks along the cell body: the head, the middle of the flagellum, and the tip of the flagellum. Target ambiguity was strongest here: the flagellum deformed more rapidly toward the tip, making the head easy to follow, the mid-flagellum harder, and the tip the most ambiguous. Together, these datasets test whether sparse human correction stays effective as the source of difficulty shifts across microscopy settings.

We compared RIPPLE-generated trajectories against exhaustive manual annotations of every frame. Benchmark datasets typically treat frame-by-frame manual annotation as the gold standard, so we used it as our reference for accuracy while also recording the human effort RIPPLE required. We quantified accuracy with average point precision (APP), the standard metric for modern computer-vision trackers \cite{doersch2022tap}, which measures how closely a predicted trajectory matches the manual comparison track across distance thresholds of 1, 2, 4, 8, and 16 pixels. Table~\ref{tab:ripple_benchmark} summarizes RIPPLE accuracy and annotation effort across the five datasets. RIPPLE achieved high APP on well-defined targets and reduced both annotation time and click counts in every dataset. As expected, datasets with visually ambiguous targets, rapid local deformation, or large temporal gaps required more user intervention. Lower APP in the most difficult datasets did not necessarily indicate where RIPPLE failed quantitatively; in all cases, RIPPLE still followed the intended biological structure, and the remaining discrepancy reflected how differently annotators placed a point on that structure.

\newlength{\rippletablewidth}
\setlength{\rippletablewidth}{\textwidth}

\newlength{\ripplethreecolwidth}
\newlength{\ripplefourcolwidth}
\newlength{\rippleeightcolwidth}

\begin{table}[t]
\centering
\small
\setlength{\tabcolsep}{4pt}
\renewcommand{\arraystretch}{1.15}
\caption{RIPPLE accuracy and annotation effort across five microscopy datasets. Manual time and clicks refer to exhaustive manual annotation. Approximate manual times ($\sim$) reflect wall-clock estimates rather than logged times; lower bounds ($>$) indicate sessions in which exhaustive manual annotation was not completed and therefore underestimate the true manual cost. Click counts are exact for both methods.}
\label{tab:ripple_benchmark}

\begin{tabular*}{\rippletablewidth}{@{\extracolsep{\fill}}lcccccc@{}}
\toprule
Dataset & Tracks & APP (\%) & \makecell{RIPPLE\\time} & \makecell{RIPPLE\\clicks} & \makecell{Manual\\time} & \makecell{Manual\\clicks} \\
\midrule
Neural \emph{Clytia} & 10 & 97.90 & 15 min & 158 & $\sim$2 h & 4,000 \\
Pinned \emph{Clytia} & 3 & 96.93 & 10 min & 116 & 27 min & 1,200 \\
Freely swimming \emph{Clytia} & 3 & 75.60 & 14 min & 352 & $>$1 h & 1,008 \\
Homeostatic \emph{Clytia} & 10 & 64.08 & 4 min & 47 & 14 min & 480 \\
Sperm QPM & 6 & 25.17 & 1 h 16 min & 529 & $>$2 h & 1,847 \\
\bottomrule
\end{tabular*}
\end{table}

\subsection{RIPPLE provides a favorable accuracy--effort tradeoff relative to automated and retraining baselines}

We next compared RIPPLE against three baselines on the Neural \emph{Clytia} dataset: TrackMate, a classical detect-and-link tracker; LocoTrack, a fully automatic deep-learning tracker from the tracking-any-point family; and SLEAP, an interactive retraining system originally developed for animal pose estimation that we adapted here for point tracking. We evaluated practical cost across four axes: manual annotations, total elapsed time, computation time, and the size of the hyper-parameter configuration space evaluated (Fig.~\ref{fig:benchmark_summary}b; Table~\ref{tab:baseline_comparison}). Note that the comparison evaluates practical effort--accuracy tradeoffs across tracking paradigms; it does not claim that retraining-based systems cannot ultimately reach higher accuracy given enough supervised training.

\begin{table}[t]
\centering
\small
\setlength{\tabcolsep}{4pt}
\renewcommand{\arraystretch}{1.15}
\caption{Accuracy and practical cost for RIPPLE and baseline trackers on the Neural \emph{Clytia} dataset. The hyper-parameter combinations column reports the size of the configuration space we evaluated to find a working setting on this dataset, not the minimum number of trials required to operate each method.}
\label{tab:baseline_comparison}

\begin{tabular*}{\rippletablewidth}{@{\extracolsep{\fill}}lccccc@{}}
\toprule
Method & APP (\%) & \makecell{Manual\\annotations} & \makecell{Total\\elapsed time} & \makecell{Computation\\time} & \makecell{Hyper-parameter\\combinations} \\
\midrule
RIPPLE    & 97.98 & 158 & 15 min     & 3.7 s     & 0 \\
TrackMate & 70.15 & 0   & 33.2 min   & 3.7 s     & 357 \\
LocoTrack & 71.18 & 0   & 2.3 s      & 2.3 s     & 0 \\
SLEAP     & 73.18 & 158 & 156.6 min  & 141.7 min & 0 \\
SLEAP-op  & 74.76 & 210 & 97.2 min   & 82.2 min  & 0 \\
\bottomrule
\end{tabular*}
\end{table}

RIPPLE achieved the highest APP while requiring only modest annotation effort, short elapsed time, 3.7~s of computation, and no parameters to tune under the reported workflow --i.e., the strongest combination of accuracy, low interaction cost, and immediate usability. TrackMate required zero manual annotations but achieved lower APP and demanded substantial dataset-specific parameter exploration. On the Neural \emph{Clytia} dataset, TrackMate did not merely miss the target by a few pixels. In this data, rapid tissue motion blurred features and shifted their brightness; out-of-focus motion dimmed targets and reduced robustness to noise. LocoTrack offered a strong fully automatic baseline, with zero annotations, 2.3~s of total elapsed time, and no parameters to tune, but also lost accuracy during rapid tissue contractions and local deformations. SLEAP, in an optimized workflow (SLEAP-op; see Methods), required more annotations, longer elapsed time, and substantial training time than RIPPLE, yet achieved lower APP than RIPPLE. Moreover, SLEAP users still had to label examples, train a model, and review outputs --each step adding to the total time. Nevertheless, we treat SLEAP not as a suboptimal baseline but as a practical example of what annotation-assisted tracking based on retraining currently looks like in applied biology. In settings where a lab is willing to invest more heavily in repeated labeling, retraining, and model refinement, SLEAP-like approaches may continue to improve until full automation is achieved. Here, the comparison shows that RIPPLE already offers a practical and favorable accuracy--effort tradeoff for trajectory construction without retraining overhead.

\subsection{RIPPLE matches its predecessor in accuracy while reducing runtime}

\begin{figure*}[t]
\centering
\includegraphics[width=0.8\textwidth]{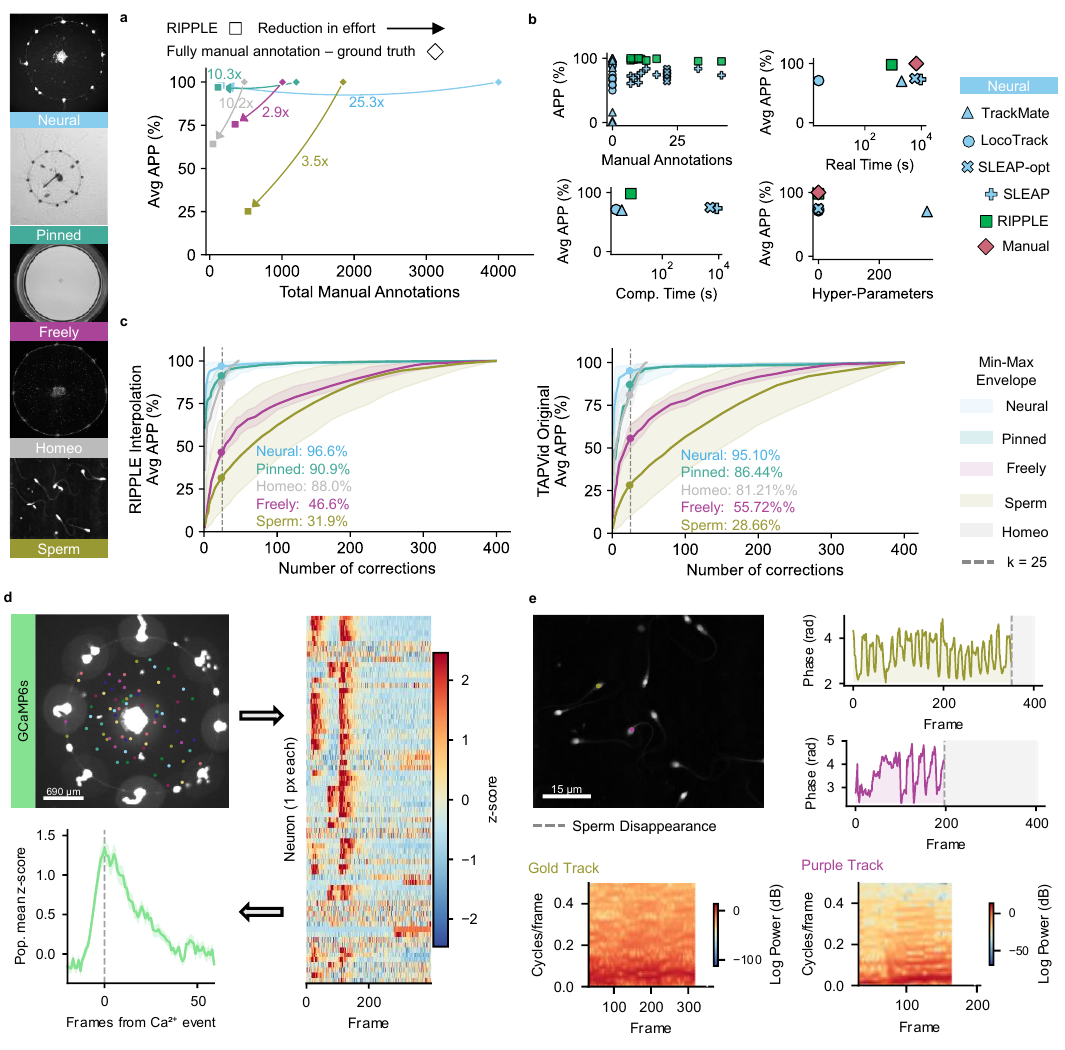}
\caption{\textbf{Benchmark summary, sparse-correction scaling, and simple downstream analyses.}
\textbf{a}, Dataset-level comparison between RIPPLE and exhaustive manual annotation, showing average point precision (APP) versus total manual insertions across the five datasets.
\textbf{b}, Neural \emph{Clytia} dataset comparison of RIPPLE, exhaustive manual annotation, TrackMate, LocoTrack, SLEAP-op, and SLEAP across four practical axes: manual annotations, total elapsed time, computation time, and the size of the hyper-parameter configuration space we evaluated to obtain a working setting (see Table~\ref{tab:baseline_comparison}). Elapsed time and computation time are shown on log-scaled x-axes. For visualization purposes, exhaustive manual annotation is included only in the elapsed-time and hyper-parameter panels.
\textbf{c}, Mean APP as a function of correction count for RIPPLE and for TAP-Vid; RIPPLE matches TAP-Vid in accuracy while providing a significant speed-up in terms of runtime
\textbf{d}, Neural \emph{Clytia} downstream analysis: single-pixel sampling of tracked GCaMP6s signals yields a neuron-by-frame activity matrix and an event-aligned population calcium average.
\textbf{e}, Sperm QPM downstream analysis: tracked phase signals yield representative phase traces and corresponding spectrograms, showing periodic structure in the sampled phase consistent with cell rolling}.
\label{fig:benchmark_summary}
\end{figure*}

To test whether RIPPLE matches the accuracy of the TAP-Vid algorithm that inspired our approach while rebuilding tracks faster, we compared the two methods directly \cite{doersch2022tap} (Fig.~\ref{fig:benchmark_summary}c; Table~\ref{tapvid}). Both methods use image motion to infer points between user-provided corrections, but they rebuild trajectories differently. We ran the publicly available TAP-Vid implementation, which uses GPU acceleration to search across the full video for the path that best agrees with the observed motion. In contrast, RIPPLE runs on a CPU, propagates positions forward and backward from neighboring anchors, and blends the two trajectories. This design allows users to run RIPPLE on a standard laptop without a GPU. Across datasets, RIPPLE and TAP-Vid produced similar APP as users added corrections, showing that RIPPLE retained the accuracy benefits of interpolating with motion. However, because we designed RIPPLE to stay lightweight, RIPPLE rebuilt tracks more than 10,000 times faster than TAP-Vid. This speed allows experts to correct trajectories interactively because the trajectory updates immediately after each new anchor.

\begin{table}[t]
\centering
\small
\setlength{\tabcolsep}{4pt}
\renewcommand{\arraystretch}{1.15}
\caption{Rebuild time and average APP at \(k=25\). Rebuild time is reported in milliseconds for each dataset and averaged across datasets.}
\label{tapvid}

\begin{tabular*}{\rippletablewidth}{@{\extracolsep{\fill}}lrrrrrrr@{}}
\toprule
 & \multicolumn{6}{c}{\makecell{Time to rebuild track at \(k=25\) (ms)}} 
 & \makecell{Avg APP\\at \(k=25\) (\%)} \\
\cmidrule(lr){2-7}
Method & Pinned & Neural & Sperm & Freely & Homeo & Mean &  \\
\midrule
RIPPLE  & 12.7   & 12.9   & 11.1   & 12.8    & 2.1     & 9.9     & 70.78 \\
TAP-Vid & 2{,}912 & 35{,}773 & 65{,}104 & 121{,}286 & 328{,}718 & 105{,}577 & 69.43 \\
\bottomrule
\end{tabular*}
\end{table}

\subsection{RIPPLE enables immediate downstream quantification}

When quantitative tracking is needed but no usable automatic tracker exists, expert-reviewed tracks still provide immediate practical value. This applies broadly to non-canonical systems such as sperm, jellyfish, amoebae, unusual cell types, plants, and other difficult targets. To demonstrate this utility, we performed two proof-of-principle quantifications on the Neural \emph{Clytia} and Sperm QPM datasets (Fig.~\ref{fig:benchmark_summary}d,e).

On the Neural \emph{Clytia} dataset, we tracked RFamide-positive neurons in a confined juvenile \emph{Clytia hemisphaerica} preparation imaged with dual-channel fluorescence microscopy, with GCaMP6s activity in the green channel. Sampling the GCaMP6s signal at the tracked neurons at single-pixel precision, produced a per-track, per-frame intensity matrix. Single-pixel sampling kept the readout sensitive to point-placement accuracy rather than to spatial averaging over a larger neighborhood. Despite preparation deformation and movement, the full-session z-score heatmap revealed two prominent events during which sampled intensity rose coherently across many tracked positions. Aligning the per-track traces to events detected from the population-mean signal yielded a clear rise near event onset followed by a slower decay. Because this signal remained coherent under single-pixel sampling, RIPPLE-generated tracks preserve point-level consistency well enough to trace neural activity.

On the Sperm dataset, we tracked sperm coordinates in quantitative phase microscopy (QPM) images. Because QPM reports optical path-length differences, phase values provide a readout of optical thickness and, under approximately fixed refractive-index contrast, can be interpreted as specimen-thickness variations \cite{park2018quantitative}. Our tracks could therefore monitor changes in the optical thickness of the sperm heads of the two sperm cells we tracked (Fig.~\ref{fig:benchmark_summary}e). Interestingly, the sperm-head thickness oscillated. A short-time spectral analysis further revealed sustained bands of elevated power rather than isolated fluctuations. This oscillation is consistent with rolling of the sperm cell around its long axis, a phenomenon previously observed through head-intensity oscillations in high-speed imaging \cite{babcock2014episodic}, dark-field/3D tracking of human sperm \cite{bukatin2015bimodal}, and optical-trap measurements of longitudinal rolling frequency and chirality \cite{zhong2022chirality}. To our knowledge, this rolling-associated signal has not previously been reported as a phase-modulation readout in QPM.

Together, these analyses show RIPPLE's ability to quantify difficult downstream tracking tasks, delivering immediate quantitative readouts in exploratory settings where a fully automatic pipeline is not yet available or reliable.

\subsection{Visual ambiguity limits APP in difficult datasets}
\label{sec:disagreement}

\begin{figure*}[t]
\centering
\includegraphics[width=\textwidth]{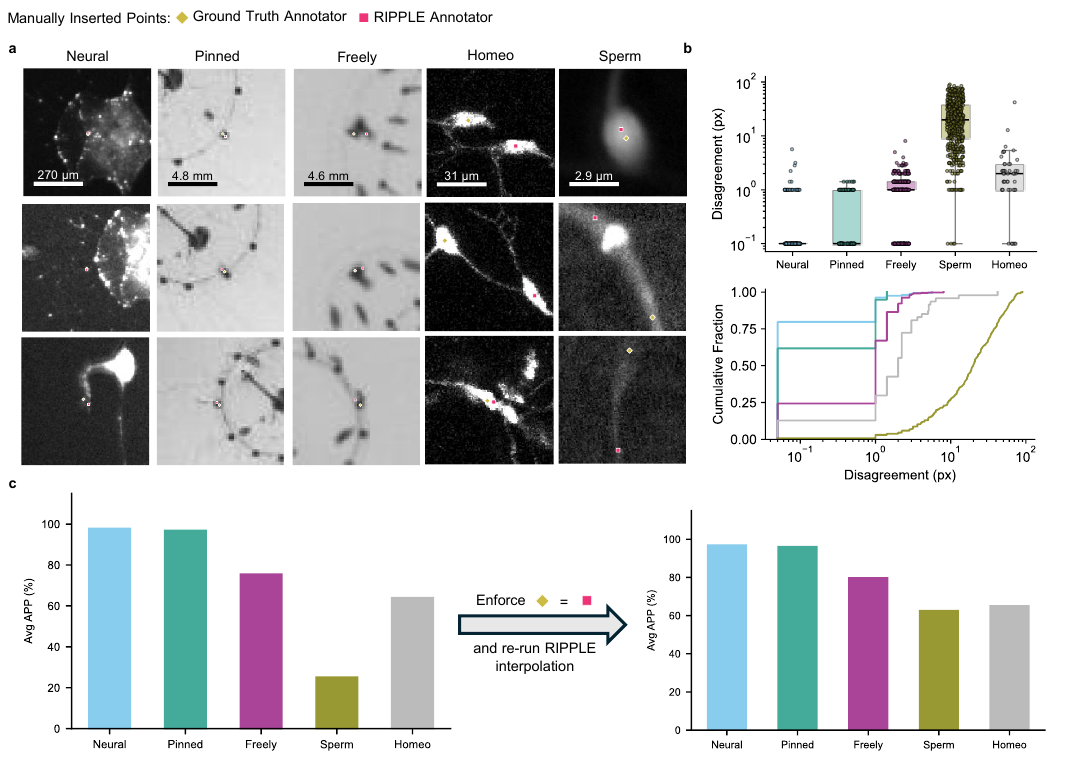}
\caption{\textbf{Annotator disagreement and its contribution to observed APP differences.}
\textbf{a}, Representative examples of matched manual insertions from the comparison annotator and the RIPPLE annotator across datasets, illustrating cases of low and high uncertainty about target placement.
\textbf{b}, Distribution and cumulative fraction of annotator disagreement measured in pixels. Neural and pinned data show near-subpixel agreement, whereas sperm exhibits substantially larger disagreement.
\textbf{c}, Analysis in which RIPPLE correction frames are preserved but the coordinates of the manually placed points are replaced with comparison coordinates before interpolation is rerun. Effects are negligible in well-defined regimes but substantial in ambiguity-limited data, consistent with annotator disagreement about target placement, amplified by APP's fixed pixel thresholds, contributing to the APP difference rather than propagation failure alone}.
\label{fig:agreement_enforcement}
\end{figure*}

Finally, we investigated why some datasets showed lower APP even when trajectories remained qualitatively consistent with the exhaustive manual comparison (Fig.~\ref{fig:agreement_enforcement}a,b). Here, we distinguished true propagation failure--where the software loses the target--from annotator disagreement about where to place a point, a common issue in quantitative bio-imaging workflows. The distinction matters for APP because its fixed distance thresholds penalize small but biologically reasonable placement differences when targets are diffuse, elongated, or hard to define precisely.

Annotators agreed at the sub-pixel level on sharp datasets such as Neural \emph{Clytia} and Pinned \emph{Clytia}, while disagreement averaged $\sim$1 pixel in Freely swimming \emph{Clytia} and $\sim$3 pixels in Homeostatic \emph{Clytia}. On Sperm QPM, disagreement was larger ($\sim$25 pixels on average) because the structure itself is elongated, making it hard to define a single exact point from the image alone. In such image data, where the pixel sampling resolution is higher than the optical resolution, each diffraction-limited feature spans many pixels causing annotations to drift even when annotators agree on the intended structure. Because APP uses fixed thresholds capped at 16~pixels, it strongly penalizes this expected divergence on visually ambiguous structures.

\section{Discussion}

\label{sec:discussion}

RIPPLE addresses a practical bottleneck in microscopy: producing accurate point trajectories in regimes where fully automatic tracking remains brittle but exhaustive manual annotation is impractical. Across five biologically distinct and technically challenging datasets, sparse motion-guided correction consistently reduced the effort to construct tracks while preserving strong agreement with exhaustive manual references. RIPPLE performed best in well-defined regimes such as Neural and Pinned \emph{Clytia}, but its value extended into harder cases involving free swimming, long temporal gaps, and visually ambiguous sperm structures. RIPPLE, therefore, is not a replacement for high-throughput automation, but an intermediate annotation infrastructure: a way to obtain trustworthy trajectories now while simultaneously producing the data needed for later benchmarking, adaptation, and model training.

The Neural \emph{Clytia} comparison illustrates this role concretely. RIPPLE reached 97.98\% APP with 158 annotations and 15 minutes of elapsed time, with no settings to tune. TrackMate reached 70.15\% APP only after substantial parameter exploration. These numbers capture the cost of obtaining TrackMate output, but they do not capture the additional effort users would spend correcting that output. To probe that correction burden, we performed a supplementary intervention-cost analysis across the matched trajectories from all five datasets. Users would need an estimated \(1.4\times\) more manual interventions to correct TrackMate output than to correct the same trajectories with RIPPLE (Supplementary Note~1; Supplementary Table~1). This estimate still favors TrackMate because it excludes the time users spend tuning parameters, filtering residual tracks, and moving between tools to inspect, correct, and reassemble fragmented trajectories. LocoTrack reached 71.18\% APP without tuning, but lost accuracy during rapid contractions and local deformations. SLEAP-op reached 74.76\% APP, but required 210 annotations, 97.2 minutes of elapsed time, and 82.2 minutes of model training. The point of these numbers is less the comparison among packages than the comparison across paradigms: TrackMate represents classical detect-and-link tracking, LocoTrack represents fully automatic tracking-any-point methods, and SLEAP provides the closest available embodiment of human-in-the-loop retraining for sparse landmark tracking. To our knowledge, no directly comparable human-in-the-loop retraining framework currently exists specifically for point tracking. We therefore included SLEAP (originally developed for pose estimation) as a deliberate paradigm representative: its interface, retraining loop, and prediction-correction workflow let users repurpose it for sparse landmark tracking, making it the strongest available proxy for what annotation-assisted retraining currently looks like in applied biology.

The five datasets also map out where sparse correction succeeds and where the target itself becomes the limiting factor. Neural and Pinned \emph{Clytia} contained compact, visually distinct targets and yielded APP values near 98\% and 97\%, respectively. Adding global translation, rotation, and self-occlusion (Freely swimming \emph{Clytia}) or large temporal gaps that broke visual continuity (Homeostatic \emph{Clytia}) increased the user's uncertainty about what to follow. Sperm QPM pushed this further: the flagellum deformed continuously, the distal end was often faint, and where exactly the tip should be placed was genuinely ambiguous compared to the head or mid-flagellum. RIPPLE reduced manual effort across all of these settings, but the reported APP fell as target ambiguity increased, which is a property of the datasets themselves rather than a single failure mode of the tool \cite{doersch2022tap}.

This pattern exposes a limit of the metric, not just of the tool. We adopted APP because it is the standard benchmark for point tracking in modern computer vision, but APP averages fixed pixel thresholds (1, 2, 4, 8, and 16~px) that do not account for target extent, image resolution, or how precisely a structure can be localized at all. A compact fluorescent neuron and a weak elongated sperm flagellum are scored on the same rigid scale. To probe how much of the apparent error in the harder datasets reflects the metric versus the tracker, we replaced the RIPPLE annotator's correction-frame coordinates with the comparison annotator's coordinates and reran the interpolation. In the sharper datasets the change was negligible (less than 1.2 percentage points in Neural, Pinned, and Homeostatic \emph{Clytia}). In Freely swimming \emph{Clytia} APP rose by approximately 4 points, and in Sperm QPM by 38 points, from 25\% to 63\%. A substantial fraction of the apparent gap therefore reflects disagreement about where on the structure to place a point, amplified by APP's fixed thresholds, rather than the tracker losing the structure itself \cite{doersch2022tap}. APP alone is therefore not a complete way to judge sparse-correction tools in regimes where the biological target is hard to define; in those regimes, an important part of RIPPLE's value is precisely that it helps experts produce coherent tracks despite that ambiguity.

That role matters most where automated methods remain brittle and exhaustive manual annotation is too expensive to scale, such as new biological systems, emerging imaging modalities, and non-canonical targets for which no reliable tracking pipeline yet exists. In those settings, RIPPLE provides quantified trajectories immediately while also lowering the cost of producing data for later benchmarking, adaptation, or training \cite{cao2025rethinking,karaev2025cotracker3,zheng2023pointodyssey}. Exploratory biology benefits in particular: pilot studies, evolving imaging setups, and one-off perturbations need quantitative feedback before it is clear that the system will justify the cost of building and tuning a dedicated tracker. RIPPLE offers a middle ground that makes quantitative tracking feasible immediately, without that overhead.

The same focus distinguishes RIPPLE from workflows that depend on retraining. Inspired in part by TAP-Vid \cite{doersch2022tap}, RIPPLE shares the use of sparse human input but differs in goal: data generation rather than model adaptation. Retraining-based workflows can produce tracks, but their primary purpose is to refit a model, which adds substantial overhead and makes them less practical in the regimes studied here. RIPPLE instead produces high-quality data efficiently---precisely the kind of data that retraining-based approaches ultimately require to yield reliable trackers. Once gold-standard tracks are available, they also support later automation: a lab can use RIPPLE to build trustworthy trajectories in a new regime, infer dynamics quickly, then use those trajectories first to benchmark existing methods and later to adapt or fine-tune newer point-tracking models. The immediate output is a set of tracks; the longer-term return is supervision for future automated inference. RIPPLE therefore fits naturally between brittle out-of-the-box tracking and reliable tracking that has already been adapted to the domain.

The current implementation has clear limits. The underlying optical-flow estimator \cite{kroeger2016fast} computes how the image shifts between consecutive frames, but it can lose reliability under severe out-of-plane motion, strong blur, signal loss, or complete target disappearance, all of which increase the need for human intervention. The sparse-correction workflow is also bounded by human attention: very large studies with tens of thousands of densely packed targets over long recordings will still demand more expert time than is practical for exhaustive review, even with substantial reductions in click count. These constraints define the scope of the tool rather than undermining its core contribution. RIPPLE is most useful when researchers need tracks for difficult new microscopy data, need them sooner than a custom pipeline would allow, and cannot rely on existing automated methods to follow the target over time.

Taken together, these results show that RIPPLE provides a practical route to point trajectories from difficult microscopy videos when automatic trackers are not yet reliable and exhaustive manual annotation is impractical. By exploiting image motion to extend each manual correction across many frames, RIPPLE reduces effort while producing tracks of high enough quality to enable immediate downstream quantification, benchmarking, and future model adaptation. It thereby fills a missing layer between manual annotation and fully automated tracking: helping researchers measure biological dynamics now, while producing the data the field needs to evaluate, adapt, and train the next generation of microscopy trackers.

\section{Methods}

\subsection{RIPPLE architecture}

We implemented RIPPLE as a standalone client--server system. The graphical frontend was written in Java (Java 17) using Swing and ImageJ \cite{collins2007imagej} for image display, TIFF loading, navigation, and overlay rendering. The computational backend was implemented in Python (Python 3.10+), which performs pixel motion estimation, propagation from the starting point, and track updating from manual correction points. The frontend and backend communicate synchronously through JSON messages separated by line breaks over a local socket connection. When a user places a starting point or inserts a manual correction point, the backend recalculates the track and triggers an immediate visual update in the interface.

\subsection{Motion estimation}

We estimated motion between consecutive frames using Dense Inverse Search (DIS) pixel motion estimation \cite{kroeger2016fast}, implemented in OpenCV \cite{bradski2000opencv}. We selected DIS after benchmarking candidate classical and deep optical-flow algorithms and comparing how many corrections each required to reach APP \(\geq 90\%\) (Supplementary Tables 1 and 2). For a sequence of length $T$ with spatial dimensions $H \times W$, the backend produces a flow tensor
\[
\mathbf{F} \in \mathbb{R}^{(T-1)\times H \times W \times 2},
\]
where $\mathbf{F}_t(y,x) = (\Delta x,\Delta y)$ denotes the displacement at pixel $(x,y)$ from frame $t$ to frame $t+1$.

\subsection{Track construction from sparse user input}

We generated trajectories in two stages. A user first places a starting point $\mathbf{p}(t_0) = (y_0,x_0)$ on frame $t_0$. RIPPLE extends the trajectory forward and backward by iterative flow lookup:
\begin{equation}
\mathbf{p}(t+1) = \mathbf{p}(t) + \mathbf{F}_t\bigl(\mathbf{p}(t)\bigr),
\qquad
\mathbf{p}(t-1) = \mathbf{p}(t) - \mathbf{F}_{t-1}\bigl(\mathbf{p}(t)\bigr).
\label{eq:methods_prop_new}
\end{equation}
Flow values at non-integer coordinates are sampled using bilinear interpolation with boundary clamping.

If the propagated trajectory drifts, the user inserts a manual correction point $(t_j,\mathbf{a}_j)$, where $t_j$ denotes the frame and $\mathbf{a}_j = (x_j,y_j)$ is the corrected position. RIPPLE reconstructs each segment between adjacent manual points by propagating the left manual point forward and the right manual point backward, producing forward and backward traces $\mathbf{f}(t)$ and $\mathbf{b}(t)$, which are then blended:
\begin{equation}
\hat{\mathbf{p}}(t) = (1-\alpha)\mathbf{f}(t) + \alpha \mathbf{b}(t),
\qquad
\alpha = \frac{t-t_0}{t_1-t_0}.
\label{eq:methods_blend_new}
\end{equation}
This guarantees that the track passes exactly through the user-defined manual points while distributing residual propagation error smoothly across the interval.

\subsection{Microscopy datasets}

The jellyfish \textit{Clytia hemisphaerica} is an emerging model for systems and behavioral neuroscience \cite{weissbourd2021genetically}. \textit{Clytia} are tiny, flat, and transparent animals that possess a mostly two-dimensional nervous system that seamlessly integrates new neurons as they grow \cite{weissbourd2021genetically,chari2021whole,houliston2010clytia}. They also offer a broad molecular and genetic toolkit, including CRISPR gene knockouts, RNAi knockdowns, targeted cell ablation, and calcium imaging \cite{weissbourd2021genetically,momose2018high,houliston2022past}. This combination of tools and morphology enables whole-animal nervous system imaging at single-cell resolution \cite{weissbourd2021genetically,houliston2022past}. To bridge scales from neuronal growth to network activity and behavior, effective tools are needed to detect and track individual neurons in live animals \cite{weissbourd2021genetically,chari2021whole}.


\subsubsection{Neural \emph{Clytia} dataset}

We generated the Neural \emph{Clytia} dataset from a transgenic line that expressed GCaMP6s and mCherry under the control of the RFamide promoter \cite{weissbourd2021genetically}. We flattened the animals in artificial seawater between a glass slide and a coverslip with Vaseline. We acquired the data on an Olympus BX51WI widefield microscope and performed dual-color imaging in the red and green channels simultaneously. We used a $4\times$ objective with a numerical aperture of 0.28. We used a custom dual-camera setup that imaged the red and green channels simultaneously; this setup consisted of two Teledyne Photometrics Prime 95B cameras and a Hamamatsu image splitter (Model A12801-10). The dataset had a pixel size of 5.48~\textmu m/pixel and image dimensions of $600 \times 600$ pixels. We acquired the data with an exposure time of 200~ms for about 1 hour. We imaged GCaMP6s in the green channel and mCherry in the red channel. We tracked the soma of a peptidergic subpopulation of neurons that expressed the neuropeptide RFamide, and we defined each tracked point as the cell body, or soma, of an RFamide neuron. We did not apply any preprocessing before analysis. The dataset contained 1 animal and multiple neurons from that same animal. During acquisition, we encountered substantial deformation, jitter, and some signal-to-noise limitations because we imaged live animals that could bleach over time.

\subsubsection{Pinned \emph{Clytia} dataset}

The Pinned \emph{Clytia} video clip was generated from a wildtype female medusa (\(\sim 6\)--8~mm in diameter) from strain Z4B \cite{leclere2019genome}. We restrained the animal by pinning it through the mouth on a Sylgard-coated plate filled with artificial seawater. Video data were acquired from above at 150~Hz using a Flea3 camera from FLIR and the manufacturer's acquisition software, FlyCapture. The single 2.67~s video clip used for analysis had image dimensions of \(100 \times 100\) pixels and contained 400 frames. The dataset contained 3 tracks: one gonad and two tentacle bulbs. During acquisition, we did not encounter major technical challenges; however, in some frames, a tentacle bulb or gonad could briefly pass behind the pin, and transient overlap between tentacle bulbs and gonads could occur during swim pulses.

\subsubsection{Freely swimming \emph{Clytia} dataset}

The Freely swimming \emph{Clytia} dataset was generated from a mature wildtype female medusa (\(\sim 10\)~mm in diameter) from strain Z4B \cite{leclere2019genome}. The animal was placed in artificial seawater in a 90~mm diameter \(\times\) 50~mm depth glass dish and imaged from above at 15~Hz using a Flea3 camera from FLIR and the acquisition software FlyCapture. The single 26.67~s video clip used for analysis had image dimensions of \(1280 \times 1024\) pixels and contained 400 frames. The dataset contained 3 tracks: the mouth, a tentacle bulb, and a gonad. During acquisition, self-occlusion could occur in freely swimming animals, and tracked structures could disappear from view when the animal swam near the wall of the dish. Reflections from the imaging setup could also affect visibility in some frames.

\subsubsection{Homeostatic \emph{Clytia} dataset}

We used a previously generated transgenic \emph{Clytia} line in which nitroreductase and mCherry were expressed under the control of the RFamide promoter \cite{weissbourd2021genetically}. We embedded \emph{Clytia} medusae in a 0.75\% low-melt agarose solution (Millipore Sigma A0701-25G) in artificial seawater, then flattened the animals and placed them in glass-bottom dishes for imaging (Pelco Glass Bottom Dishes, 50 $\times$ 6~mm, Ted Pella Inc. 14035-20). We acquired the data on a Zeiss LSM 900 confocal microscope and collected tiled z-stacks at 20-minute intervals. We used a $10\times$ objective with a numerical aperture of 0.45. We detected the signal with the LSM confocal detector. The dataset had a pixel size of 0.623~\textmu m/pixel and image dimensions of $5632 \times 5632$ pixels. We collected 48 frames over 16 hours at 20-minute intervals. We acquired the images in the red channel and imaged mCherry. We tracked the soma of a peptidergic subpopulation of neurons that expressed the neuropeptide RFamide, and we defined each tracked point as the cell body, or soma, of an RFamide neuron. We did not apply any preprocessing before analysis. The dataset contained 1 animal and multiple neurons from that same animal. During acquisition, we encountered substantial deformation, jitter, and some signal-to-noise limitations because we imaged live animals that could bleach over time.



\subsection{Downstream quantification}

\subsubsection{Neural calcium analysis}

For the Neural \emph{Clytia} dataset, we used tracks we produced with RIPPLE to sample the green-channel GCaMP6s signal at each tracked neuronal position. Using RIPPLE, we produced 71 tracks, yielding $n=71$ neuronal sampling locations. At each frame, we sampled the pixel intensity at the tracked $(x,y)$ coordinate.

We then computed the baseline-normalized fluorescence change for each neuron as
\[
\Delta F/F_0 = \frac{F(t)-F_0}{F_0},
\]
where \(F(t)\) is the sampled fluorescence intensity at frame \(t\), and \(F_0\) is the mean intensity over the first 11 frames of the trace, corresponding to the quiescent period preceding the earliest detected population activity in this recording. We then z-scored each neuron's full \(\Delta F/F_0\) trace as
\[
z(t) = \frac{\Delta F/F_0(t)-\mu}{\sigma},
\]
where \(\mu\) and \(\sigma\) denote the mean and standard deviation of that neuron's full \(\Delta F/F_0\) trace. To identify candidate coordinated calcium events, we averaged the z-scored traces across all 71 neurons at each frame and detected peaks in this population-mean trace using scipy.signal.find\_peaks with height = 0.8, prominence = 0.5, and minimum inter-peak distance = 15 frames. These parameters were chosen empirically to isolate prominent population transients in this recording. We then extracted peri-event windows spanning 20 frames before to 60 frames after each detected event. For each neuron, we averaged its peri-event responses across detected events, then summarized the result as the population mean \(\pm\) SEM across neurons. We displayed the full neuron-by-frame z-score matrix as a heatmap and ordered neurons by the frame index of their peak z-score response.

\subsubsection{Sperm phase analysis}

For the Sperm QPM dataset, we generated two representative tracks with RIPPLE for downstream analysis. At each frame, we sampled the phase value at the tracked \((x, y)\) coordinate. We used the annotation metadata generated with RIPPLE to identify the first frame at which the target moved out of sight, and we truncated each trace at that disappearance onset so that downstream analysis used only visible segments.

We computed a short-time Fourier spectrogram using scipy.signal.spectrogram, reported spectral power in decibels, and expressed frequency in cycles per frame. Time axes are reported in frames.

\subsection{Manual annotation}

We used exhaustive manual annotation as the reference standard for evaluation in all datasets. For each trajectory, an annotator specified the target position in every scored frame using a consistent, dataset-specific target heuristic. We treat these annotations as expert manual references rather than absolute physical ground truth.

To reduce arbitrary pixel-level variability, we adjusted the annotation cursor or brush size to approximately match the visible extent of the tracked feature and used the centroid of the marked region as the exact comparison point. During RIPPLE annotation, the user inserted only the number of manual correction points needed to produce a satisfactory trajectory. Exhaustive manual annotations were generated using 3D Slicer \cite{kikinis20133d}. When an occlusion temporarily hid the target, the annotator moved forward in the video to re-identify it and then traced it backward through the ambiguous interval as consistently as possible. In some cases---particularly in the freely swimming \emph{Clytia} dataset---self-occlusion near the boundary of the visible field made it impossible to recover the intended target.

\subsection{Comparator methods}

We compared RIPPLE against TrackMate \cite{tinevez2017trackmate}, LocoTrack \cite{cho2024local}, and SLEAP \cite{pereira2022sleap}. These tools were selected because they represent classical detection-and-linking, learned long-range point matching, and based on retraining user-guided tracking, respectively. We applied the comparator methods to the Neural \emph{Clytia} dataset and used this dataset for the practical comparison shown in Fig.~\ref{fig:benchmark_summary}b. For TrackMate, we manually tested settings for this dataset and selected the ones that best recovered the intended targets while reducing spurious detections and linking errors. To measure how much tuning TrackMate required, we counted each time we paired one tested detector setting with one tested linker setting as one candidate setup. We tested 17 detector settings: DoG detector (\(n=4\)), Hessian detector (\(n=5\)), label image detector (\(n=1\)), LoG detector (\(n=4\)), mask detector (\(n=1\)), and thresholding detector (\(n=2\)). We also tested 21 linker settings: Simple LAP tracker (\(n=3\)), LAP tracker (\(n=6\)), Advanced Kalman tracker (\(n=5\)), Kalman tracker (\(n=3\)), Overlap tracker (\(n=3\)), and Nearest Neighbor tracker (\(n=1\)). Because we paired each detector setting with each linker setting, TrackMate required us to consider \(17 \times 21 = 357\) possible setups. We did not count manual annotation or manual tracking as automated TrackMate settings. In the final TrackMate setup, we used the DoG detector to identify candidate points, which took 1~s, and the Advanced Kalman Tracking algorithm to link them across frames, which took 2.7~s. For LocoTrack, we deployed the base model and supplied all query points jointly. Because SLEAP asks users to label examples, collect those labels, and retrain the model, we tested it under two supervision setups. First, to approximate how RIPPLE handles each track, we annotated and evaluated one track at a time. For each track, we opened a separate SLEAP session, placed the same annotations that RIPPLE used, trained the model with a 90:10 training-validation split, and inferred the intermediate frames. Second, in the SLEAP-op setup, we inserted corrections every 20 frames across all tracks to better match how users typically apply SLEAP. We then trained the model with the same 90:10 split and inferred the frames between corrections. We evaluated all methods against the same manual comparison trajectories.

\subsection{Tracking accuracy metric}

We quantified tracking accuracy using the TAP-Vid point-tracking metric \cite{doersch2022tap}, reported as average point precision (APP). APP measures agreement with the manual comparison trajectory at distance thresholds of 1, 2, 4, 8, and 16 pixels. For a given threshold $\tau$, point precision is defined as
\[
\mathrm{PP}(\tau)
=
\Pr\!\left(
\left\lVert \hat{\mathbf{p}} - \mathbf{p} \right\rVert_2 \le \tau
\mid \text{point visible}
\right),
\]
or equivalently in discrete time as
\[
\mathrm{PP}(\tau)
=
\frac{\sum_t v_t \,\mathbf{1}\!\left[
\left\lVert \hat{\mathbf{p}}(t)-\mathbf{p}(t) \right\rVert_2 \le \tau
\right]}
{\sum_t v_t},
\]
where $\mathbf{p}(t)$ is the manual comparison coordinate, $\hat{\mathbf{p}}(t)$ is the predicted coordinate, and $v_t$ indicates whether the point was visible and scored at frame $t$. Following the TAP-Vid standard, we evaluated $\tau \in \{1,2,4,8,16\}$ and averaged across thresholds to obtain a single APP value per trajectory:
\[
\mathrm{APP}
=
\frac{1}{5}
\sum_{\delta \in \{1,2,4,8,16\}}
\mathrm{APP}(\delta).
\]
Trajectory-level APP values were then averaged to produce dataset-level APP scores.

Because APP relies on fixed pixel thresholds capped at 16~px, it does not account for landmark extent, image resolution, or human annotator uncertainty. This approximation works well in compact, visually sharp datasets. However, in visually ambiguous datasets, differences in APP can reflect a combination of genuine tracking failures, human disagreement, and mismatch between the rigid 16-px threshold and the physical size or visual extent of the tracked feature. We therefore treat APP as an operational benchmark rather than a complete measure of tracking quality in all biological contexts.

\subsection{Manual effort and timing measurements}

We recorded two primary efficiency metrics: manual effort and elapsed time. Manual effort was defined as the total number of manual point insertions made by the user. For timing, we recorded both pure processing time and total wall-clock session time, including preprocessing. We report both because they capture complementary aspects of real-world cost. Annotation count provides a strict, objective comparison across workflows, whereas elapsed time also reflects visual inspection, decision-making, and software interaction overhead. Elapsed time naturally varies across annotators, and fully characterizing that variability would require a larger user study.

\subsection{Sparse-correction scaling analysis}

To understand how tracking performance scales with human effort, we simulated RIPPLE’s correction process. Starting from the initial point alone, we progressively inserted manual comparison positions as correction points and evaluated the resulting trajectory after each insertion. For each track, we added correction points in random order across 100 independent trials and measured APP after every addition. For each correction count, we selected the trial that produced the highest accuracy. We repeated this experiment for all tracks across all datasets, averaged APP across tracks at each correction count, and plotted the mean together with the min–max envelope.

\subsection{TAP-Vid optical-flow control}

To test whether RIPPLE preserves the accuracy benefits of motion-based interpolation while reducing rebuild time, we compared RIPPLE with the publicly released optical-flow track-assist implementation from the official TAPNet repository, which accompanies TAP-Vid \cite{doersch2022tap,tapnet_optical_flow_track_assist}. This implementation uses dense optical flow to infer point positions between user-provided anchors. In the released notebook, TAPNet computes dense optical flow with RAFT and uses dynamic programming to find a path between two consecutive anchors that best agrees with the estimated frame-to-frame motion. The notebook notes that this public implementation differs from the original TAP-Vid annotation system, which used Dijkstra's algorithm.

Both methods received the same anchors at each value of \(k\). For each \(k\), we generated multiple random orders in which corrections entered the trajectory and rebuilt the track with both methods. We then evaluated APP for each rebuilt trajectory and reported the best APP achieved by each method at that \(k\). This procedure removes dependence on one specific correction order and focuses the comparison on how each algorithm rebuilds trajectories from the same sparse anchors.

\subsection{Annotator disagreement and manual point-replacement analysis}

To quantify how consistently annotators defined targets across datasets, we measured the Euclidean distance (in pixels) between corresponding points placed independently by the comparison annotator and the RIPPLE annotator on the same frames. We then summarized these disagreement distributions separately for each dataset.

To estimate how much disagreement about target placement contributed to APP differences, we designed a manual point-replacement experiment. We took the exact frames where the user placed RIPPLE corrections, replaced the RIPPLE correction-point coordinates with the exhaustive comparison coordinates, and reran the interpolation procedure. We then compared the newly generated trajectories to the manual comparison track using APP. This experiment isolates the effect of human point-placement disagreement while holding correction-point timing and the interpolation algorithm constant.

\subsection{Software dependencies}

We developed the RIPPLE frontend in Java using Swing and ImageJ \cite{collins2007imagej}. The backend was implemented in Python using OpenCV for pixel motion estimation \cite{bradski2000opencv}, NumPy for array operations, SciPy for interpolation, and Pillow and tifffile for image I/O. Software environments were managed using Maven for the Java frontend and Conda for the Python backend. The software exports finalized trajectories as JSON files containing per-frame coordinates and relevant track metadata. The complete Python dependency list for the CPU and GPU environments is provided in Supplementary Table 3.

\begin{backmatter}

\bmsection{Acknowledgements}
K.C. is a Shurl and Kay Curci Awardee of the Life Sciences Research Foundation. J.O.K. is a Picower Postdoctoral Fellow, funded by The Picower Institute for Learning and Memory and The Freedom Together Foundation. B.W. acknowledges support from The Kavli Foundation, NSF Award \#2449561, The Picower Institute for Learning and Memory, and The Freedom Together Foundation. We thank Mona Nystad for providing the sperm samples used in this study. We also thank Dr. Frank Y. Liu for discussions that helped shape the project’s conceptual direction.

\bmsection{Author contributions}

L.Z. and D.N.W. conceived the study. L.Z. designed the optical-flow interpolation algorithm, implemented the software, developed the experimental framework, performed the experiments, annotated data, generated the figures and visualizations, and wrote the initial manuscript. P.T. contributed to software design, study planning, and data annotation. K.B. performed extensive data annotation and helped develop the annotation heuristics. D.J. contributed to software design and study planning. J.O.K. generated the Neural \emph{Clytia} and Homeostatic \emph{Clytia} datasets, and K.C. generated the Pinned \emph{Clytia} datasets. J.O.K. and X.L. contributed to data annotation and extensively tested the software, providing user-centered feedback that improved it. A.A. generated the sperm QPM dataset. B.S.A. supervised A.A. and provided conceptual guidance for the sperm-data component and its downstream use. N.C. provided feedback that helped shape the study. B.W. provided conceptual guidance, defined the downstream tasks, and generated the freely swimming Clytia dataset. D.N.W. supervised the work, advised L.Z., and led the project. All authors revised the manuscript and approved the final version.

\bmsection{Competing Interests}

The authors declare no competing interests.

\bmsection{Data availability}

The data underlying the results in this study are not publicly available at this time because of ongoing project considerations, but may be made available from the corresponding authors upon request.

\bmsection{Code and supplementary resources}

The RIPPLE software and source code are publicly available at
\url{https://github.com/Le0nZim/ripple}. The GitHub repository also provides access to all the supplementary resources, including the demo videos, uncompressed supplementary movies, LocoTrack failure-mode videos, and SLEAP performance videos. The interactive demo can also be accessed directly at
\url{https://huggingface.co/spaces/Le0nZim/ripple-demo}.

\end{backmatter}

\putbib

\end{bibunit}

\clearpage

\begin{bibunit}

\setcounter{figure}{0}
\setcounter{table}{0}

\renewcommand{\thefigure}{S\arabic{figure}}
\renewcommand{\thetable}{S\arabic{table}}

\renewcommand{\figurename}{Supplementary Figure}
\renewcommand{\tablename}{Supplementary Table}

\title{Motion-guided sparse correction enables expert-quality point tracking across diverse microscopy regimes}

\author{Leonidas Zimianitis,\authormark{1}
Pasindu Thenahandi,\authormark{1,$\dag$}
Kai Buckhalter,\authormark{1,$\dag$}
Dineth Jayakody,\authormark{1}
Julian O. Kimura,\authormark{2,3}
Xinyue Liang,\authormark{2,3}
Karen Cunningham,\authormark{2,3}
Azeem Ahmad,\authormark{4}
Balpreet S. Ahluwalia,\authormark{4,5}
Sampath Jayarathna,\authormark{1}
Nikos Chrisochoides,\authormark{1,6}
Brandon Weissbourd,\authormark{2,3}
and Dushan N. Wadduwage\authormark{1,6,7,*}}

\address{\authormark{1}Department of Computer Science, Old Dominion University, Norfolk, VA 23529, USA\\
\authormark{2}Department of Biology, Massachusetts Institute of Technology, Cambridge, MA 02139, USA\\
\authormark{3}The Picower Institute for Learning and Memory, Massachusetts Institute of Technology, Cambridge, MA 02139, USA\\
\authormark{4}Department of Physics and Technology, UiT--The Arctic University of Norway, Troms\o{} 9037, Norway\\
\authormark{5}Department of Physics, University of Oslo, Oslo 0316, Norway\\
\authormark{6}School of Data Science, Old Dominion University, Norfolk, VA 23529, USA\\
\authormark{7}Department of Physics, Old Dominion University, Norfolk, VA 23529, USA\\
\authormark{$\dag$}These authors contributed equally.}

\email{\authormark{*}dwadduwa@odu.edu}

\section*{Supplementary Material}

Supplementary Figures 1-2 \\
Supplementary Tables 1-3 \\
Supplementary Experiment


\begin{figure*}[t]
    \centering
    \includegraphics[width=\textwidth]{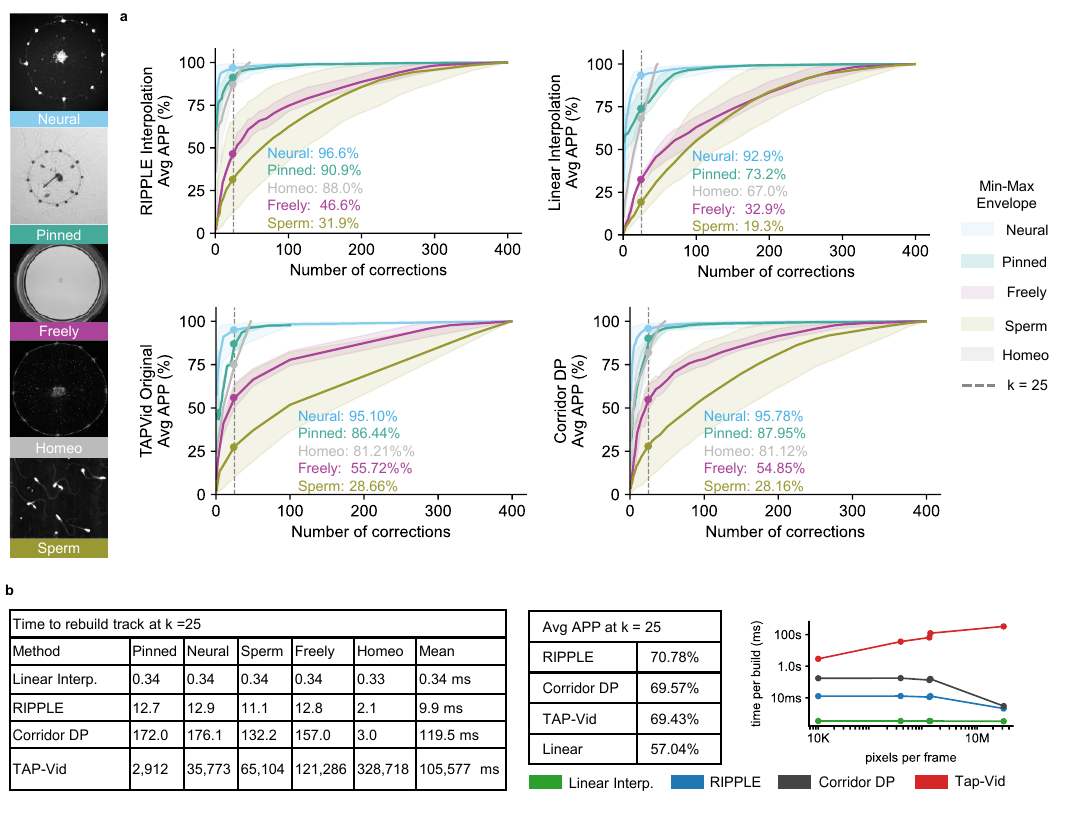}
    \caption{\textbf{Comparison of interpolation strategies, accuracy, and computational cost.}
    \textbf{a}, Accuracy as a function of the number of manual correction points for four interpolation strategies across the five microscopy datasets. RIPPLE flow-blend interpolation uses dense image-motion estimates to propagate sparse user corrections, whereas linear interpolation connects the same correction points with straight-line trajectories. TAP-Vid original denotes the original dynamic-programming interpolation procedure from the TAP-Vid framework, and Corridor-DP denotes our custom corridor-based dynamic-programming implementation based on TAP-Vid. Curves show mean average point precision (APP) across tracks, and shaded regions indicate the min--max envelope. The dashed vertical line marks \(k=25\) manual corrections. At this sparse-correction operating point, RIPPLE flow-blend interpolation achieved higher APP than linear interpolation and performed similarly to dynamic-programming-based interpolation, while requiring substantially less computation than the original TAP-Vid implementation.
    \textbf{b}, Runtime and accuracy summary at \(k=25\). Linear interpolation was fastest but produced lower APP. RIPPLE flow-blend interpolation achieved the highest average APP at \(k=25\) with millisecond-scale rebuild times, whereas the TAP-Vid original implementation required substantially longer rebuild times. The implementation improvements produced an \(883\times\) speedup from TAP-Vid original to Corridor-DP and an additional \(12.1\times\) speedup from Corridor-DP to RIPPLE flow-blend interpolation, making RIPPLE interpolation \(10{,}684\times\) faster than TAP-Vid original overall. Linear interpolation provided a further \(29.1\times\) speedup relative to RIPPLE flow-blend interpolation, but at the cost of reduced accuracy in the sparse-correction regime. Runtime scaling with image size shows that RIPPLE and linear interpolation remain fast as the number of pixels per frame increases, while TAP-Vid original becomes orders of magnitude slower.}
    \label{fig:supp_interp_comparison}
\end{figure*}

\clearpage

\begin{figure*}[t]
    \centering
    \includegraphics[width=\textwidth]{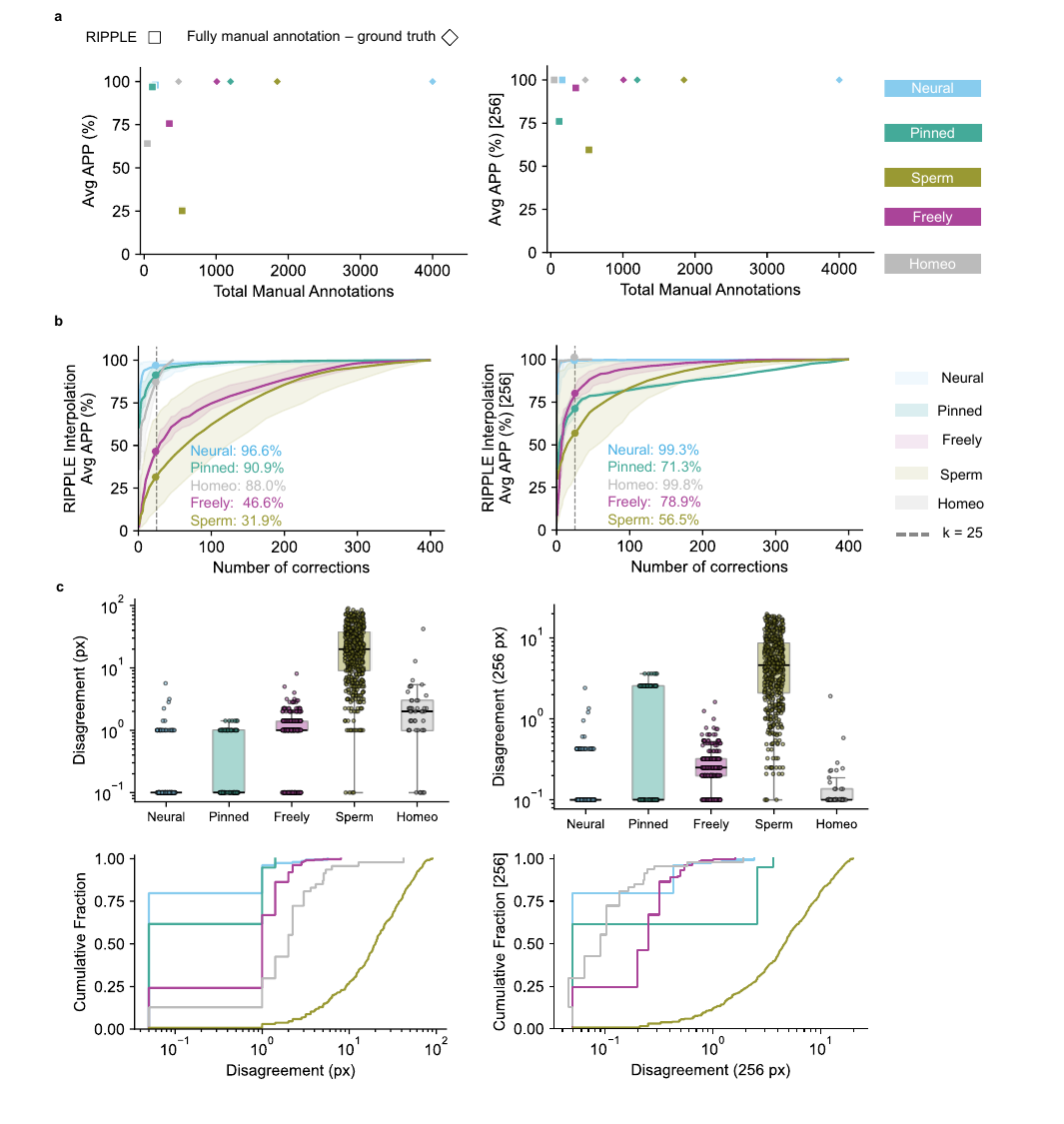}
    \caption{\textbf{a}, Comparison between RIPPLE sparse annotation and exhaustive manual annotation before and after rescaling all datasets to 256 \(\times\) 256 pixels, the resolution used in the TAP-Vid evaluation setting from which the APP metric was adopted. Each point shows dataset-level APP versus the total number of manual annotations. Diamonds indicate fully manual annotation, used as the comparison reference, and squares indicate RIPPLE-generated trajectories. Rescaling changes the effective pixel scale used by the fixed APP thresholds. \textbf{b}, RIPPLE interpolation APP as a function of correction count in the original image resolution and after rescaling to 256 \(\times\) 256 pixels. Curves show mean APP and shaded regions show the min--max envelope. Datasets originally larger than 256 \(\times\) 256 generally show increased APP after rescaling because the fixed 1-, 2-, 4-, 8-, and 16-pixel APP thresholds become less stringent relative to the original field of view. In contrast, the pinned \emph{Clytia} dataset decreases after rescaling because its native resolution is 100 \(\times\) 100 pixels, so rescaling to 256 \(\times\) 256 makes the same physical errors occupy more pixels. \textbf{c}, Annotator disagreement before and after 256 \(\times\) 256 rescaling. Box plots and cumulative distributions show that apparent disagreement changes with image scale, emphasizing that fixed pixel-threshold metrics such as APP depend on resolution as well as tracking quality.}
    \label{fig:supp_resolution_scaling}
\end{figure*}

\clearpage

\setlength{\rippletablewidth}{\textwidth}

{
\rippletableformat
\setlength{\LTleft}{0pt}
\setlength{\LTright}{0pt}
\setlength{\ripplethreecolwidth}{\dimexpr\rippletablewidth-4\tabcolsep\relax}

\begin{longtable}{@{}L{0.18\ripplethreecolwidth} L{0.25\ripplethreecolwidth} L{0.57\ripplethreecolwidth}@{}}
    \caption{Candidate backbone optical flow algorithms benchmarked for RIPPLE.}
    \label{tab:candidate-backbone-flow-algorithms} \\

    \hline
    \textbf{Package} & \textbf{Algorithm} & \textbf{Description} \\
    \hline
    \endfirsthead

    \hline
    \textbf{Package} & \textbf{Algorithm} & \textbf{Description} \\
    \hline
    \endhead

    \hline
    \endfoot

    opencv & dis\_ultrafast & DIS (Dense Inverse Search) --- ultrafast preset~\cite{kroeger2016fast} \\
    opencv & dis\_fast & DIS --- fast preset~\cite{kroeger2016fast} \\
    opencv & dis\_medium & DIS --- medium preset (default in RIPPLE)~\cite{kroeger2016fast} \\
    opencv & farneback & Gunnar Farneb{\"a}ck's polynomial expansion method~\cite{farneback2003two} \\
    \hline

    opencv\_contrib & dual\_tvl1 & Dual TV-L1 variational method~\cite{zach2007duality} \\
    opencv\_contrib & dense\_rlof & Dense Robust Local Optical Flow~\cite{senst2012robust} \\
    opencv\_contrib & deepflow & DeepFlow --- large displacement matching + variational~\cite{weinzaepfel2013deepflow} \\
    opencv\_contrib & simpleflow & SimpleFlow --- fast dense flow~\cite{tao2012simpleflow} \\
    opencv\_contrib & pcaflow & PCA-Flow --- PCA-based prior on flow fields~\cite{wulff2015efficient} \\
    opencv\_contrib & sparse\_to\_dense & Sparse-to-Dense interpolation flow~\cite{bradski2000opencv} \\
    \hline

    scikit-image & skimage\_ilk & Iterative Lucas-Kanade~\cite{le2005dense} \\
    scikit-image & skimage\_tvl1 & TV-L1 variational method~\cite{zach2007duality} \\
    \hline

    pyoptflow & horn\_schunck & Horn-Schunck global variational method~\cite{horn1981determining} \\
    \hline

    torchvision & tv\_raft\_large & RAFT large~\cite{teed2020raft} \\
    torchvision & tv\_raft\_small & RAFT small~\cite{teed2020raft} \\
    \hline

    ptlflow & ptl\_raft & RAFT~\cite{teed2020raft} \\
    ptlflow & ptl\_raft\_small & RAFT small~\cite{teed2020raft} \\
    ptlflow & ptl\_gma & GMA --- Global Motion Aggregation~\cite{jiang2021learning} \\
    ptlflow & ptl\_flowformer & FlowFormer --- transformer-based optical flow~\cite{huang2022flowformer} \\
    ptlflow & ptl\_flowformer\_pp & FlowFormer++~\cite{shi2023flowformer++} \\
    ptlflow & ptl\_gmflow & GMFlow --- global matching flow~\cite{xu2022gmflow} \\
    ptlflow & ptl\_unimatch & UniMatch / GMFlow+~\cite{xu2023unifying} \\
    ptlflow & ptl\_sea\_raft\_l & SEA-RAFT large~\cite{wang2024sea} \\
    ptlflow & ptl\_ms\_raft\_p & MS-RAFT+ multi-scale~\cite{jahedi2022high} \\
    ptlflow & ptl\_neuflow2 & NeuFlow v2 --- efficient neural flow~\cite{zhang2025neuflow} \\
    ptlflow & ptl\_ccmr & CCMR --- cost volume cross-attention~\cite{jahedi2024ccmr} \\
    ptlflow & ptl\_craft & CRAFT --- cross-attentional flow~\cite{sui2022craft} \\
    ptlflow & ptl\_memflow & MemFlow --- memory-based flow~\cite{dong2024memflow} \\
    ptlflow & ptl\_skflow & SKFlow --- selective kernel flow~\cite{sun2022skflow} \\
    ptlflow & ptl\_rpknet & RPKNet --- recurrent prediction kernel~\cite{morimitsu2024recurrent} \\
    ptlflow & ptl\_rapidflow & RAPIDFlow --- efficient recurrent flow~\cite{morimitsu2024rapidflow} \\
    ptlflow & ptl\_videoflow\_bof & VideoFlow --- bidirectional optical flow~\cite{shi2023videoflow} \\
    ptlflow & ptl\_pwcnet & PWC-Net~\cite{sun2018pwc} \\
    ptlflow & ptl\_liteflownet & LiteFlowNet~\cite{hui2018liteflownet} \\
    ptlflow & ptl\_maskflownet\_s & MaskFlownet-S~\cite{zhao2020maskflownet} \\
    ptlflow & ptl\_flownets & FlowNetS~\cite{dosovitskiy2015flownet} \\
    ptlflow & ptl\_flownet2 & FlowNet 2.0~\cite{ilg2017flownet} \\
    ptlflow & ptl\_irr\_pwc & IRR-PWC~\cite{hur2019iterative} \\
    ptlflow & ptl\_separableflow & SeparableFlow~\cite{zhang2021separable} \\
    ptlflow & ptl\_starflow & STaRFlow~\cite{godet2021starflow} \\
    ptlflow & ptl\_dip & DIP --- deep inverse patchmatch~\cite{zheng2022dip} \\
    ptlflow & ptl\_matchflow & MatchFlow~\cite{dong2023rethinking} \\
    ptlflow & ptl\_scopeflow & ScopeFlow~\cite{bar2020scopeflow} \\
    ptlflow & ptl\_streamflow & StreamFlow --- streaming flow~\cite{sun2024streamflow} \\
    ptlflow & ptl\_dpflow & DPFlow --- dual-pyramid flow~\cite{morimitsu2025dpflow} \\
    ptlflow & ptl\_flow1d & Flow1D --- 1D attention flow~\cite{xu2021high} \\
    ptlflow & ptl\_fastflownet & FastFlowNet --- lightweight fast flow~\cite{kong2021fastflownet} \\
    ptlflow & ptl\_csflow & CSFlow --- cross-strip flow~\cite{shi2022csflow} \\
    ptlflow & ptl\_llaflow & LLA-Flow --- local-level attention~\cite{xu2023lla} \\
    ptlflow & ptl\_gmflownet & GMFlowNet --- global matching + FlowNet~\cite{zhao2022global} \\
    ptlflow & ptl\_dicl & DICL --- displacement-invariant cost learning~\cite{wang2020displacement} \\
    \hline
\end{longtable}
}

{
\rippletableformat
\setlength{\LTleft}{0pt}
\setlength{\LTright}{0pt}
\setlength{\rippleeightcolwidth}{\dimexpr\rippletablewidth-14\tabcolsep\relax}

\begin{longtable}{@{}C{0.05\rippleeightcolwidth} L{0.29\rippleeightcolwidth} L{0.18\rippleeightcolwidth} C{0.096\rippleeightcolwidth} C{0.096\rippleeightcolwidth} C{0.096\rippleeightcolwidth} C{0.096\rippleeightcolwidth} C{0.096\rippleeightcolwidth}@{}}
    \caption{Correction annotations needed to reach APP $\geq 90\%$ for candidate backbone optical flow algorithms benchmarked for RIPPLE.}
    \label{tab:flow-benchmark-app90-no-regen} \\

    \hline
    \textbf{\#} & \textbf{Algorithm} & \textbf{Package} & \textbf{Pinned} & \textbf{Neural} & \textbf{Sperm} & \textbf{Freely} & \textbf{Avg} \\
    \hline
    \endfirsthead

    \hline
    \textbf{\#} & \textbf{Algorithm} & \textbf{Package} & \textbf{Pinned} & \textbf{Neural} & \textbf{Sperm} & \textbf{Freely} & \textbf{Avg} \\
    \hline
    \endhead

    \hline
    \endfoot

    1 & dense\_rlof & opencv\_contrib & 6 & 35 & --- & --- & - \\
    2 & simpleflow & opencv\_contrib & 40 & 5 & --- & 250 & - \\
    3 & skimage\_tvl1 & scikit-image & 2 & 1 & 200 & 250 & 113 \\
    4 & farneback & opencv & 1 & 1 & 237 & 225 & 116 \\
    5 & skimage\_ilk & scikit-image & 1 & 1 & 237 & 225 & 116 \\
    6 & dis\_medium & opencv & 25 & 2 & 212 & 225 & 116 \\
    7 & dis\_fast & opencv & 40 & 3 & 200 & 225 & 117 \\
    8 & dis\_ultrafast & opencv & 35 & 2 & 212 & 225 & 119 \\
    9 & dual\_tvl1 & opencv\_contrib & 25 & 2 & 200 & 250 & 119 \\
    10 & deepflow & opencv\_contrib & 35 & 3 & 200 & 250 & 122 \\
    11 & pcaflow & opencv\_contrib & 30 & 5 & 212 & 250 & 124 \\
    12 & ptl\_dpflow & ptlflow & 35 & 6 & 212 & 250 & 126 \\
    13 & ptl\_dip & ptlflow & 40 & 5 & 212 & 250 & 127 \\
    14 & sparse\_to\_dense & opencv\_contrib & 30 & 1 & 225 & 250 & 127 \\
    15 & ptl\_rapidflow & ptlflow & 50 & 6 & 212 & 250 & 130 \\
    16 & ptl\_rpknet & ptlflow & 50 & 6 & 212 & 250 & 130 \\
    17 & horn\_schunck & pyoptflow & 40 & 10 & 225 & 250 & 131 \\
    18 & ptl\_ccmr & ptlflow & 40 & 12 & 225 & 250 & 132 \\
    19 & ptl\_ms\_raft\_p & ptlflow & 45 & 7 & 225 & 250 & 132 \\
    20 & ptl\_gmflow & ptlflow & 60 & 7 & 212 & 250 & 132 \\
    21 & ptl\_unimatch & ptlflow & 60 & 7 & 212 & 250 & 132 \\
    22 & ptl\_videoflow\_bof & ptlflow & 45 & 35 & 237 & 250 & 142 \\
    23 & tv\_raft\_large & torchvision & --- & 9 & 212 & 250 & - \\
    24 & tv\_raft\_small & torchvision & --- & 9 & 212 & 250 & - \\
    25 & ptl\_flowformer & ptlflow & --- & --- & 212 & 250 & - \\
    26 & ptl\_flowformer\_pp & ptlflow & --- & --- & 212 & 250 & - \\
    27 & ptl\_raft\_small & ptlflow & 300 & 350 & 275 & 275 & 300 \\
    28 & ptl\_flow1d & ptlflow & 350 & 337 & 287 & 300 & 319 \\
    29 & ptl\_dicl & ptlflow & --- & 375 & 287 & 300 & - \\
    30 & ptl\_flownets & ptlflow & 350 & 350 & 287 & 300 & 322 \\
    31 & ptl\_raft & ptlflow & 350 & 350 & 287 & 300 & 322 \\
    32 & ptl\_csflow & ptlflow & 375 & 375 & 287 & 300 & 334 \\
    33 & ptl\_fastflownet & ptlflow & 375 & 375 & 287 & 300 & 334 \\
    34 & ptl\_flownet2 & ptlflow & 375 & 375 & 287 & 300 & 334 \\
    35 & ptl\_gma & ptlflow & 375 & 375 & 287 & 300 & 334 \\
    36 & ptl\_gmflownet & ptlflow & 375 & 375 & 287 & 300 & 334 \\
    37 & ptl\_irr\_pwc & ptlflow & 375 & 375 & 287 & 300 & 334 \\
    38 & ptl\_liteflownet & ptlflow & 375 & 375 & 287 & 300 & 334 \\
    39 & ptl\_llaflow & ptlflow & 375 & 375 & 287 & 300 & 334 \\
    40 & ptl\_maskflownet\_s & ptlflow & 375 & 375 & 287 & 300 & 334 \\
    41 & ptl\_memflow & ptlflow & 375 & 375 & 287 & 300 & 334 \\
    42 & ptl\_neuflow2 & ptlflow & 375 & 375 & 287 & 300 & 334 \\
    43 & ptl\_pwcnet & ptlflow & 375 & 375 & 287 & 300 & 334 \\
    44 & ptl\_scopeflow & ptlflow & 375 & 375 & 287 & 300 & 334 \\
    45 & ptl\_sea\_raft\_l & ptlflow & 375 & 375 & 287 & 300 & 334 \\
    46 & ptl\_skflow & ptlflow & 375 & 375 & 287 & 300 & 334 \\
    47 & ptl\_starflow & ptlflow & 375 & 375 & 287 & 300 & 334 \\
    48 & ptl\_craft & ptlflow & 350 & 375 & --- & --- & - \\
    \hline
\end{longtable}
}

\begin{table}[ht]
    \centering
    \caption{Python packages used in RIPPLE. The ``Mode'' column indicates whether the package is installed in the CPU build, the GPU (CUDA) build, or both.}
    \label{tab:ripple-packages}
    \small
    \setlength{\tabcolsep}{3pt}
    \renewcommand{\arraystretch}{1.08}
    \setlength{\ripplefourcolwidth}{\dimexpr\rippletablewidth-6\tabcolsep\relax}

    \begin{tabular}{@{}L{0.24\ripplefourcolwidth} L{0.40\ripplefourcolwidth} L{0.20\ripplefourcolwidth} L{0.16\ripplefourcolwidth}@{}}
        \hline
        \textbf{Category} & \textbf{Package} & \textbf{Version} & \textbf{Mode} \\
        \hline
        \multirow{3}{*}{Scientific computing}
            & numpy                          & $\geq$2.0.0     & Both \\
            & scipy                          & $\geq$1.10.0    & Both \\
            & pandas                         & $\geq$2.0.0     & Both \\
        \hline
        \multirow{6}{*}{Image processing}
            & tifffile                       & $\geq$2023.2.0  & Both \\
            & imageio                        & $\geq$2.28.0    & Both \\
            & mediapy                        & $\geq$1.2.0     & Both \\
            & opencv-contrib-python-headless & $\geq$4.7.0     & Both \\
            & scikit-image                   & $\geq$0.20.0    & Both \\
            & pillow                         & $\geq$10.0.0    & Both \\
        \hline
        \multirow{3}{*}{PyTorch}
            & torch                          & $\geq$2.0.0     & Both \\
            & torchvision                    & $\geq$0.15.0    & Both \\
            & torchaudio                     & $\geq$2.0.0     & GPU \\
        \hline
        \multirow{4}{*}{PyTorch ecosystem}
            & pytorch-lightning              & $\geq$2.0.0     & GPU \\
            & lightning                      & $\geq$2.0.0     & GPU \\
            & torchmetrics                   & $\geq$1.0.0     & GPU \\
            & einops                         & $\geq$0.6.0     & GPU \\
        \hline
        \multirow{5}{*}{TensorFlow (LocoTrack)}
            & tensorflow                     & $\geq$2.15.0    & GPU \\
            & tensorflow-addons              & $\geq$0.23.0    & GPU \\
            & tensorflow-datasets            & $\geq$4.9.0     & GPU \\
            & tensorflow-graphics            & $\geq$2021.12.0 & GPU \\
            & keras                          & $\geq$3.0.0     & GPU \\
        \hline
        Particle tracking
            & trackpy                        & $\geq$0.6.0     & Both \\
        \hline
        \multirow{2}{*}{Data handling}
            & h5py                           & $\geq$3.8.0     & Both \\
            & pyarrow                        & $\geq$12.0.0    & GPU \\
        \hline
        \multirow{3}{*}{System utilities}
            & psutil                         & $\geq$5.9.0     & Both \\
            & tqdm                           & $\geq$4.65.0    & Both \\
            & requests                       & $\geq$2.28.0    & Both \\
        \hline
        Visualization
            & matplotlib                     & $\geq$3.7.0     & Both \\
        \hline
        \multirow{3}{*}{Utilities}
            & pyyaml                         & $\geq$6.0       & Both \\
            & click                          & $\geq$8.0.0     & Both \\
            & rich                           & $\geq$13.0.0    & Both \\
        \hline
    \end{tabular}
\end{table}

\section*{Supplementary Experiment: Estimating TrackMate's manual intervention cost for benchmark trajectories}

We estimated how much manual frame-to-frame intervention a user would need to reconstruct the selected benchmark trajectories from two starting points:

\begin{itemize}
  \item \textbf{RIPPLE}: the anchors placed by the human annotator during the RIPPLE session for each matched track.
  \item \textbf{TrackMate} (v7.14.0): the unfiltered output of the classical detect-and-link pipeline, exported with the TrackMate XML exporter.
\end{itemize}

\subsection*{Selecting the tracks of interest}

We selected the trajectories by matching the manual segmentations to the RIPPLE annotations. For each dataset, we first computed the centre of mass of each manual segmentation mask at each frame to obtain a ground-truth trajectory in image pixel coordinates. We then compared each ground-truth trajectory with each RIPPLE trajectory in the same dataset using their positions at frame~0, or the first visible frame when a trajectory did not appear at frame~0. We greedily paired each ground-truth trajectory with the unused RIPPLE trajectory whose first visible position lay closest to it. This matching procedure produced $32$ ground-truth trajectories, which we used as the tracks of interest for the intervention-cost analysis.

\subsection*{Counting RIPPLE interventions}

For each matched RIPPLE trajectory, we counted the number of anchors that the annotator placed. Each anchor contains a frame index and an $(x,y)$ position. We summed the number of anchors across all matched tracks in each dataset. This count directly reports the number of clicks that the annotator performed during the RIPPLE session.

\subsection*{Counting TrackMate interventions}

For each matched ground-truth trajectory, we compared the ground-truth position against the TrackMate output frame by frame. We counted three types of user intervention:

\begin{itemize}
  \item init. pick: one click to select the correct TrackMate trajectory at the first visible frame of a ground-truth track.
  \item relink: one click when the closest valid TrackMate detection belonged to a different TrackMate fragment than the fragment the user had been following.
  \item manual one click: when no TrackMate detection fell within the matching tolerance of the ground-truth position.
\end{itemize}

For each ground-truth trajectory, we initialized the followed TrackMate fragment as undefined. At each visible frame, we found the TrackMate detection closest to the ground-truth position, provided that the distance fell below the same tolerance used during track matching. When no detection satisfied this criterion, we counted one manual intervention and reset the followed fragment. When the followed fragment was undefined, we counted one init\_pick intervention and set the followed fragment to the current TrackMate fragment. When the current detection belonged to a different TrackMate fragment, we counted one relink intervention and updated the followed fragment. Otherwise, we counted no intervention.

We defined the total TrackMate intervention cost as
init\_pick+relink+manual,
summed over all matched tracks. This estimate gives TrackMate a favourable lower bound: it treats every relink and every manual placement as a single frictionless click, and it assumes that the user can identify the correct target position at each frame.

Two TrackMate exports stored coordinates in physical units rather than pixels and did not include a calibration block. For those cases, we converted the detections to image pixels using an isotropic scale estimated from the largest observed coordinate and the corresponding image dimension. This heuristic gave an approximate pixel calibration for comparing TrackMate detections with the ground-truth trajectories.

\subsection*{Results}

\begin{table}[h]
\centering
\small
\setlength{\tabcolsep}{6pt}
\renewcommand{\arraystretch}{1.15}

\begin{tabular*}{\rippletablewidth}{@{\extracolsep{\fill}}lrrrrr@{}}
\toprule
\textbf{Dataset} & \textbf{Tracks} & \textbf{GT frames}
  & \textbf{TrackMate} & \textbf{RIPPLE} & \textbf{TM/RIP}\\
\midrule
Neural Activity & 10 & \num{4000} & \num{251}  & \num{158}  & $1.6\times$\\
Pinned Down     &  3 & \num{1199} & \textbf{\num{27}}  & \num{116}  & $0.2\times$\\
Freely Swimming &  3 & \num{1008} & \textbf{\num{267}} & \num{352}  & $0.8\times$\\
Sperm           &  6 & \num{1847} & \num{1007} & \num{529}  & $1.9\times$\\
Homeostasis     & 10 & \num{480}  & \num{90}   & \num{47}   & $1.9\times$\\
\midrule
\textbf{All}    & \textbf{32} & \textbf{\num{8534}}
                & \textbf{\num{1642}} & \textbf{\num{1202}}
                & $\mathbf{1.4\times}$\\
\bottomrule
\end{tabular*}

\caption{Manual interventions required to reconstruct the selected benchmark trajectories from each starting point. TrackMate interventions include the initial trajectory selection, relinking events, and fully manual point placements. RIPPLE interventions correspond to the anchors placed by the annotator. TM/RIP reports how many times more interventions TrackMate required than RIPPLE.}
\label{tab:ripple-vs-trackmate}
\end{table}

\subsection*{Summary}

Across all 32 matched trajectories, TrackMate required \num{1642} estimated interventions, whereas RIPPLE required \num{1202} measured interventions. Thus, RIPPLE reduced the overall intervention count by approximately $1.4\times$, even though the TrackMate estimate used an optimistic lower-bound counting scheme. This estimate counts only the interventions needed after obtaining the TrackMate output. It does not include the time required to tune TrackMate parameters, filter the output to retain the relevant tracks, or inspect and reject spurious fragments. In practice, more permissive detector settings can recover more true targets but also produce more residual tracks, which increases the burden of finding the correct fragment to relink. The estimate also does not include the workflow delay introduced by exporting, inspecting, correcting, and reassembling fragmented TrackMate trajectories outside the initial tracking step. RIPPLE required fewer interventions on the more challenging datasets, including Neural Activity, Sperm, and Homeostasis. TrackMate required fewer interventions only on the Pinned Down dataset, where the targets moved little and appeared with high contrast. These results suggest that RIPPLE reduces manual intervention most strongly when target motion, deformation, overlap, or weak localization make classical detect-and-link tracking harder to correct.

\renewcommand{\refname}{Supplementary References}
\putbib

\end{bibunit}


\begin{thebibliography}{10}
\newcommand{\enquote}[1]{``#1''}

\bibitem{wu2022multiscale}
Y.~Wu and H.~Shroff, \enquote{Multiscale fluorescence imaging of living
  samples,} {\protect\JournalTitle{Histochemistry and cell biology}}
  \textbf{158}, 301--323 (2022).

\bibitem{pylvanainen2023live}
J.~W. Pylv{\"a}n{\"a}inen, E.~G{\'o}mez-de Mariscal, R.~Henriques, and
  G.~Jacquemet, \enquote{Live-cell imaging in the deep learning era,}
  {\protect\JournalTitle{Current Opinion in Cell Biology}} \textbf{85}, 102271
  (2023).

\bibitem{jaqaman2008robust}
K.~Jaqaman, D.~Loerke, M.~Mettlen, \emph{et~al.}, \enquote{Robust
  single-particle tracking in live-cell time-lapse sequences,}
  {\protect\JournalTitle{Nature methods}} \textbf{5}, 695--702 (2008).

\bibitem{chenouard2014objective}
N.~Chenouard, I.~Smal, F.~De~Chaumont, \emph{et~al.}, \enquote{Objective
  comparison of particle tracking methods,} {\protect\JournalTitle{Nature
  methods}} \textbf{11}, 281--289 (2014).

\bibitem{sbalzarini2005feature}
I.~F. Sbalzarini and P.~Koumoutsakos, \enquote{Feature point tracking and
  trajectory analysis for video imaging in cell biology,}
  {\protect\JournalTitle{Journal of structural biology}} \textbf{151}, 182--195
  (2005).

\bibitem{cheng2022review}
H.-J. Cheng, C.-H. Hsu, C.-L. Hung, and C.-Y. Lin, \enquote{A review for cell
  and particle tracking on microscopy images using algorithms and deep learning
  technologies,} {\protect\JournalTitle{biomedical journal}} \textbf{45},
  465--471 (2022).

\bibitem{manzo2015review}
C.~Manzo and M.~F. Garcia-Parajo, \enquote{A review of progress in single
  particle tracking: from methods to biophysical insights,}
  {\protect\JournalTitle{Reports on progress in physics}} \textbf{78}, 124601
  (2015).

\bibitem{worth2009live}
D.~C. Worth and M.~Parsons, \enquote{Live cell imaging analysis of receptor
  function,} in \emph{Live Cell Imaging: Methods and Protocols,}  (Springer,
  2009), pp. 311--323.

\bibitem{zhang2021intracellular}
M.-L. Zhang, H.-Y. Ti, P.-Y. Wang, and H.~Li, \enquote{Intracellular transport
  dynamics revealed by single-particle tracking,}
  {\protect\JournalTitle{Biophysics Reports}} \textbf{7}, 413 (2021).

\bibitem{ker2018phase}
D.~F.~E. Ker, S.~Eom, S.~Sanami, \emph{et~al.}, \enquote{Phase contrast
  time-lapse microscopy datasets with automated and manual cell tracking
  annotations,} {\protect\JournalTitle{Scientific data}} \textbf{5}, 180237
  (2018).

\bibitem{lugagne2020delta}
J.-B. Lugagne, H.~Lin, and M.~J. Dunlop, \enquote{Delta: Automated cell
  segmentation, tracking, and lineage reconstruction using deep learning,}
  {\protect\JournalTitle{PLoS computational biology}} \textbf{16}, e1007673
  (2020).

\bibitem{versari2017long}
C.~Versari, S.~Stoma, K.~Batmanov, \emph{et~al.}, \enquote{Long-term tracking
  of budding yeast cells in brightfield microscopy: Cellstar and the evaluation
  platform,} {\protect\JournalTitle{Journal of The Royal Society Interface}}
  \textbf{14} (2017).

\bibitem{li2008cell}
K.~Li, E.~D. Miller, M.~Chen, \emph{et~al.}, \enquote{Cell population tracking
  and lineage construction with spatiotemporal context,}
  {\protect\JournalTitle{Medical image analysis}} \textbf{12}, 546--566 (2008).

\bibitem{malin2023automated}
C.~Malin-Mayor, P.~Hirsch, L.~Guignard, \emph{et~al.}, \enquote{Automated
  reconstruction of whole-embryo cell lineages by learning from sparse
  annotations,} {\protect\JournalTitle{Nature biotechnology}} \textbf{41},
  44--49 (2023).

\bibitem{atanas2026deep}
A.~A. Atanas, A.~K.-Y. Lu, B.~Goodell, \emph{et~al.}, \enquote{Deep neural
  networks to register and annotate cells in moving and deforming nervous
  systems,} {\protect\JournalTitle{eLife}} \textbf{14}, RP108159 (2026).

\bibitem{cao2025rethinking}
J.~Cao, J.~Wenzel, S.~Zhang, \emph{et~al.}, \enquote{Rethinking deep learning
  in bioimaging through a data centric lens,} {\protect\JournalTitle{npj
  Imaging}} \textbf{3}, 29 (2025).

\bibitem{tinevez2017trackmate}
J.-Y. Tinevez, N.~Perry, J.~Schindelin, \emph{et~al.}, \enquote{Trackmate: An
  open and extensible platform for single-particle tracking,}
  {\protect\JournalTitle{Methods}} \textbf{115}, 80--90 (2017).

\bibitem{ershov2022trackmate}
D.~Ershov, M.-S. Phan, J.~W. Pylv{\"a}n{\"a}inen, \emph{et~al.},
  \enquote{Trackmate 7: integrating state-of-the-art segmentation algorithms
  into tracking pipelines,} {\protect\JournalTitle{Nature methods}}
  \textbf{19}, 829--832 (2022).

\bibitem{roudot2023u}
P.~Roudot, W.~R. Legant, Q.~Zou, \emph{et~al.}, \enquote{u-track3d: Measuring,
  navigating, and validating dense particle trajectories in three dimensions,}
  {\protect\JournalTitle{Cell Reports Methods}} \textbf{3} (2023).

\bibitem{kok2020organoidtracker}
R.~N. Kok, L.~Hebert, G.~Huelsz-Prince, \emph{et~al.},
  \enquote{Organoidtracker: Efficient cell tracking using machine learning and
  manual error correction,} {\protect\JournalTitle{PLoS One}} \textbf{15},
  e0240802 (2020).

\bibitem{ulman2017objective}
V.~Ulman, M.~Ma{\v{s}}ka, K.~E.~G. Magnusson, \emph{et~al.}, \enquote{An
  objective comparison of cell-tracking algorithms,}
  {\protect\JournalTitle{Nature methods}} \textbf{14}, 1141--1152 (2017).

\bibitem{stringer2021cellpose}
C.~Stringer, T.~Wang, M.~Michaelos, and M.~Pachitariu, \enquote{Cellpose: a
  generalist algorithm for cellular segmentation,}
  {\protect\JournalTitle{Nature methods}} \textbf{18}, 100--106 (2021).

\bibitem{schmidt2018cell}
U.~Schmidt, M.~Weigert, C.~Broaddus, and G.~Myers, \enquote{Cell detection with
  star-convex polygons,} in \emph{International conference on medical image
  computing and computer-assisted intervention,}  (Springer, 2018), pp.
  265--273.

\bibitem{pereira2022sleap}
T.~D. Pereira, N.~Tabris, A.~Matsliah, \emph{et~al.}, \enquote{Sleap: A deep
  learning system for multi-animal pose tracking,}
  {\protect\JournalTitle{Nature methods}} \textbf{19}, 486--495 (2022).

\bibitem{nath2019using}
T.~Nath, A.~Mathis, A.~C. Chen, \emph{et~al.}, \enquote{Using deeplabcut for 3d
  markerless pose estimation across species and behaviors,}
  {\protect\JournalTitle{Nature protocols}} \textbf{14}, 2152--2176 (2019).

\bibitem{doersch2023tapir}
C.~Doersch, Y.~Yang, M.~Vecerik, \emph{et~al.}, \enquote{{TAPIR}: Tracking any
  point with per-frame initialization and temporal refinement,} in
  \emph{Proceedings of the IEEE/CVF International Conference on Computer Vision
  (ICCV),}  (2023), pp. 10061--10072.

\bibitem{karaev2024cotracker}
N.~Karaev, I.~Rocco, B.~Graham, \emph{et~al.}, \enquote{Cotracker: It is better
  to track together,} in \emph{European conference on computer vision,}
  (Springer, 2024), pp. 18--35.

\bibitem{cho2024local}
S.~Cho, J.~Huang, J.~Nam, \emph{et~al.}, \enquote{Local all-pair correspondence
  for point tracking,} in \emph{European Conference on Computer Vision (ECCV),}
   (2024).

\bibitem{harley2022particle}
A.~W. Harley, Z.~Fang, and K.~Fragkiadaki, \enquote{Particle video revisited:
  Tracking through occlusions using point trajectories,} in \emph{ECCV,}
  (2022).

\bibitem{karaev2025cotracker3}
N.~Karaev, Y.~Makarov, J.~Wang, \emph{et~al.}, \enquote{Cotracker3: Simpler and
  better point tracking by pseudo-labelling real videos,} in \emph{Proceedings
  of the IEEE/CVF International Conference on Computer Vision,}  (2025), pp.
  6013--6022.

\bibitem{spilger2021deep}
R.~Spilger, J.-Y. Lee, V.~O. Chagin, \emph{et~al.}, \enquote{Deep probabilistic
  tracking of particles in fluorescence microscopy images,}
  {\protect\JournalTitle{Medical image analysis}} \textbf{72}, 102128 (2021).

\bibitem{hansen2018robust}
A.~S. Hansen, M.~Woringer, J.~B. Grimm, \emph{et~al.}, \enquote{Robust
  model-based analysis of single-particle tracking experiments with spot-on,}
  {\protect\JournalTitle{Elife}} \textbf{7}, e33125 (2018).

\bibitem{mavska2023cell}
M.~Ma{\v{s}}ka, V.~Ulman, P.~Delgado-Rodriguez, \emph{et~al.}, \enquote{The
  cell tracking challenge: 10 years of objective benchmarking,}
  {\protect\JournalTitle{Nature Methods}} \textbf{20}, 1010--1020 (2023).

\bibitem{zheng2023pointodyssey}
Y.~Zheng, A.~W. Harley, B.~Shen, \emph{et~al.}, \enquote{Pointodyssey: A
  large-scale synthetic dataset for long-term point tracking,} in
  \emph{Proceedings of the IEEE/CVF International Conference on Computer
  Vision,}  (2023), pp. 19855--19865.

\bibitem{van2021biological}
D.~van~der Wal, I.~Jhun, I.~Laklouk, \emph{et~al.}, \enquote{Biological data
  annotation via a human-augmenting ai-based labeling system,}
  {\protect\JournalTitle{NPJ digital medicine}} \textbf{4}, 145 (2021).

\bibitem{mavska2014benchmark}
M.~Ma{\v{s}}ka, V.~Ulman, D.~Svoboda, \emph{et~al.}, \enquote{A benchmark for
  comparison of cell tracking algorithms,}
  {\protect\JournalTitle{Bioinformatics}} \textbf{30}, 1609--1617 (2014).

\bibitem{sugawara2022tracking}
K.~Sugawara, {\c{C}}.~{\c{C}}evrim, and M.~Averof, \enquote{Tracking cell
  lineages in 3d by incremental deep learning,} {\protect\JournalTitle{Elife}}
  \textbf{11}, e69380 (2022).

\bibitem{han2019edetect}
H.~Han, G.~Wu, Y.~Li, and Z.~Zi, \enquote{edetect: a fast error detection and
  correction tool for live cell imaging data analysis,}
  {\protect\JournalTitle{Iscience}} \textbf{13}, 1--8 (2019).

\bibitem{wagner2021tracurate}
S.~Wagner, K.~Thierbach, T.~Zerjatke, \emph{et~al.}, \enquote{Tracurate:
  efficiently curating cell tracks,} {\protect\JournalTitle{SoftwareX}}
  \textbf{13}, 100656 (2021).

\bibitem{padovani2022segmentation}
F.~Padovani, B.~Mairh{\"o}rmann, P.~Falter-Braun, \emph{et~al.},
  \enquote{Segmentation, tracking and cell cycle analysis of live-cell imaging
  data with cell-acdc,} {\protect\JournalTitle{BMC biology}} \textbf{20}, 174
  (2022).

\bibitem{fukai2023laptrack}
Y.~T. Fukai and K.~Kawaguchi, \enquote{Laptrack: linear assignment particle
  tracking with tunable metrics,} {\protect\JournalTitle{Bioinformatics}}
  \textbf{39}, btac799 (2023).

\bibitem{fazeli2020automated}
E.~Fazeli, N.~H. Roy, G.~Follain, \emph{et~al.}, \enquote{Automated cell
  tracking using stardist and trackmate,}
  {\protect\JournalTitle{F1000Research}} \textbf{9}, 1279 (2020).

\bibitem{weissbourd2021genetically}
B.~Weissbourd, T.~Momose, A.~Nair, \emph{et~al.}, \enquote{A genetically
  tractable jellyfish model for systems and evolutionary neuroscience,}
  {\protect\JournalTitle{Cell}} \textbf{184}, 5854--5868 (2021).

\bibitem{chari2021whole}
T.~Chari, B.~Weissbourd, J.~Gehring, \emph{et~al.}, \enquote{Whole-animal
  multiplexed single-cell rna-seq reveals transcriptional shifts across clytia
  medusa cell types,} {\protect\JournalTitle{Science Advances}} \textbf{7},
  eabh1683 (2021).

\bibitem{houliston2022past}
E.~Houliston, L.~Lecl{\`e}re, C.~Munro, \emph{et~al.}, \enquote{Past, present
  and future of clytia hemisphaerica as a laboratory jellyfish,} in
  \emph{Current topics in developmental biology,}  vol. 147 (Elsevier, 2022),
  pp. 121--151.

\bibitem{doersch2022tap}
C.~Doersch, A.~Gupta, L.~Markeeva, \emph{et~al.}, \enquote{Tap-vid: A benchmark
  for tracking any point in a video,} {\protect\JournalTitle{Advances in Neural
  Information Processing Systems}} \textbf{35}, 13610--13626 (2022).

\bibitem{kroeger2016fast}
T.~Kroeger, R.~Timofte, D.~Dai, and L.~Van~Gool, \enquote{Fast optical flow
  using dense inverse search,} in \emph{European conference on computer
  vision,}  (Springer, 2016), pp. 471--488.

\bibitem{park2018quantitative}
Y.~Park, C.~Depeursinge, and G.~Popescu, \enquote{Quantitative phase imaging in
  biomedicine,} {\protect\JournalTitle{Nature photonics}} \textbf{12}, 578--589
  (2018).

\bibitem{babcock2014episodic}
D.~F. Babcock, P.~M. Wandernoth, and G.~Wennemuth, \enquote{Episodic rolling
  and transient attachments create diversity in sperm swimming behavior,}
  {\protect\JournalTitle{BMC biology}} \textbf{12}, 67 (2014).

\bibitem{bukatin2015bimodal}
A.~Bukatin, I.~Kukhtevich, N.~Stoop, \emph{et~al.}, \enquote{Bimodal rheotactic
  behavior reflects flagellar beat asymmetry in human sperm cells,}
  {\protect\JournalTitle{Proceedings of the National Academy of Sciences}}
  \textbf{112}, 15904--15909 (2015).

\bibitem{zhong2022chirality}
Z.~Zhong, C.~Zhang, R.~Liu, \emph{et~al.}, \enquote{Chirality and frequency
  measurement of longitudinal rolling of human sperm using optical trap,}
  {\protect\JournalTitle{Frontiers in Bioengineering and Biotechnology}}
  \textbf{10}, 1028857 (2022).

\bibitem{collins2007imagej}
T.~J. Collins, \enquote{Imagej for microscopy,}
  {\protect\JournalTitle{Biotechniques}} \textbf{43}, S25--S30 (2007).

\bibitem{bradski2000opencv}
G.~Bradski, A.~Kaehler \emph{et~al.}, \enquote{Opencv,}
  {\protect\JournalTitle{Dr. Dobb's journal of software tools}} \textbf{3},
  1--81 (2000).

\bibitem{houliston2010clytia}
E.~Houliston, T.~Momose, and M.~Manuel, \enquote{Clytia hemisphaerica: a
  jellyfish cousin joins the laboratory,} {\protect\JournalTitle{Trends in
  Genetics}} \textbf{26}, 159--167 (2010).

\bibitem{momose2018high}
T.~Momose, A.~De~Cian, K.~Shiba, \emph{et~al.}, \enquote{High doses of
  crispr/cas9 ribonucleoprotein efficiently induce gene knockout with low
  mosaicism in the hydrozoan clytia hemisphaerica through
  microhomology-mediated deletion,} {\protect\JournalTitle{Scientific Reports}}
  \textbf{8}, 11734 (2018).

\bibitem{leclere2019genome}
L.~Lecl{\`e}re, C.~Horin, S.~Chevalier, \emph{et~al.}, \enquote{The genome of
  the jellyfish clytia hemisphaerica and the evolution of the cnidarian
  life-cycle,} {\protect\JournalTitle{Nature ecology \& evolution}} \textbf{3},
  801--810 (2019).

\bibitem{kikinis20133d}
R.~Kikinis, S.~D. Pieper, and K.~G. Vosburgh, \enquote{3d slicer: a platform
  for subject-specific image analysis, visualization, and clinical support,} in
  \emph{Intraoperative imaging and image-guided therapy,}  (Springer, 2013),
  pp. 277--289.

\bibitem{tapnet_optical_flow_track_assist}
{Google DeepMind}, \enquote{{Demo for Annotating a Point Track with Optical
  Flow},}
  \url{https://github.com/google-deepmind/tapnet/blob/main/colabs/optical_flow_track_assist.ipynb}
  (2026). Official TAPNet repository. Accessed: 2026-04-26.

\end{thebibliography}


\begin{thebibliography}{10}
\newcommand{\enquote}[1]{``#1''}

\bibitem{kroeger2016fast}
T.~Kroeger, R.~Timofte, D.~Dai, and L.~Van~Gool, \enquote{Fast optical flow
  using dense inverse search,} in \emph{European conference on computer
  vision,}  (Springer, 2016), pp. 471--488.

\bibitem{farneback2003two}
G.~Farneb{\"a}ck, \enquote{Two-frame motion estimation based on polynomial
  expansion,} in \emph{Scandinavian conference on Image analysis,}  (Springer,
  2003), pp. 363--370.

\bibitem{zach2007duality}
C.~Zach, T.~Pock, and H.~Bischof, \enquote{A duality based approach for
  realtime tv-l 1 optical flow,} in \emph{Joint pattern recognition symposium,}
   (Springer, 2007), pp. 214--223.

\bibitem{senst2012robust}
T.~Senst, V.~Eiselein, and T.~Sikora, \enquote{Robust local optical flow for
  feature tracking,} {\protect\JournalTitle{IEEE Transactions on Circuits and
  Systems for Video Technology}} \textbf{22}, 1377--1387 (2012).

\bibitem{weinzaepfel2013deepflow}
P.~Weinzaepfel, J.~Revaud, Z.~Harchaoui, and C.~Schmid, \enquote{Deepflow:
  Large displacement optical flow with deep matching,} in \emph{Proceedings of
  the IEEE international conference on computer vision,}  (2013), pp.
  1385--1392.

\bibitem{tao2012simpleflow}
M.~Tao, J.~Bai, P.~Kohli, and S.~Paris, \enquote{Simpleflow: A non-iterative,
  sublinear optical flow algorithm,} {\protect\JournalTitle{Computer Graphics
  Forum}} \textbf{31}, 345--353 (2012).

\bibitem{wulff2015efficient}
J.~Wulff and M.~J. Black, \enquote{Efficient sparse-to-dense optical flow
  estimation using a learned basis and layers,} in \emph{Proceedings of the
  IEEE Conference on Computer Vision and Pattern Recognition,}  (2015), pp.
  120--130.

\bibitem{bradski2000opencv}
G.~Bradski, A.~Kaehler \emph{et~al.}, \enquote{Opencv,}
  {\protect\JournalTitle{Dr. Dobb's journal of software tools}} \textbf{3},
  1--81 (2000).

\bibitem{le2005dense}
G.~Le~Besnerais and F.~Champagnat, \enquote{Dense optical flow by iterative
  local window registration,} in \emph{IEEE International Conference on Image
  Processing 2005,}  vol.~1 (IEEE, 2005), pp. I--137.

\bibitem{horn1981determining}
B.~K. Horn and B.~G. Schunck, \enquote{Determining optical flow,}
  {\protect\JournalTitle{Artificial intelligence}} \textbf{17}, 185--203
  (1981).

\bibitem{teed2020raft}
Z.~Teed and J.~Deng, \enquote{{RAFT}: Recurrent all-pairs field transforms for
  optical flow,} in \emph{European Conference on Computer Vision (ECCV),}
  (Springer, 2020), pp. 402--419.

\bibitem{jiang2021learning}
S.~Jiang, D.~Campbell, Y.~Lu, \emph{et~al.}, \enquote{Learning to estimate
  hidden motions with global motion aggregation,} in \emph{Proceedings of the
  IEEE/CVF international conference on computer vision,}  (2021), pp.
  9772--9781.

\bibitem{huang2022flowformer}
Z.~Huang, X.~Shi, C.~Zhang, \emph{et~al.}, \enquote{Flowformer: A transformer
  architecture for optical flow,} in \emph{European conference on computer
  vision,}  (Springer, 2022), pp. 668--685.

\bibitem{shi2023flowformer++}
X.~Shi, Z.~Huang, D.~Li, \emph{et~al.}, \enquote{Flowformer++: Masked cost
  volume autoencoding for pretraining optical flow estimation,} in
  \emph{Proceedings of the IEEE/CVF conference on computer vision and pattern
  recognition,}  (2023), pp. 1599--1610.

\bibitem{xu2022gmflow}
H.~Xu, J.~Zhang, J.~Cai, \emph{et~al.}, \enquote{Gmflow: Learning optical flow
  via global matching,} in \emph{Proceedings of the IEEE/CVF conference on
  computer vision and pattern recognition,}  (2022), pp. 8121--8130.

\bibitem{xu2023unifying}
H.~Xu, J.~Zhang, J.~Cai, \emph{et~al.}, \enquote{Unifying flow, stereo and
  depth estimation,} {\protect\JournalTitle{IEEE Transactions on Pattern
  Analysis and Machine Intelligence}} \textbf{45}, 13941--13958 (2023).

\bibitem{wang2024sea}
Y.~Wang, L.~Lipson, and J.~Deng, \enquote{Sea-raft: Simple, efficient, accurate
  raft for optical flow,} in \emph{European Conference on Computer Vision,}
  (Springer, 2024), pp. 36--54.

\bibitem{jahedi2022high}
A.~Jahedi, M.~Luz, L.~Mehl, \emph{et~al.}, \enquote{High resolution multi-scale
  raft (robust vision challenge 2022),} {\protect\JournalTitle{arXiv preprint
  arXiv:2210.16900}}  (2022).

\bibitem{zhang2025neuflow}
Z.~Zhang, A.~Gupta, H.~Jiang, and H.~Singh, \enquote{Neuflow-v2: Push
  high-efficiency optical flow to the limit,} in \emph{2025 IEEE/RSJ
  International Conference on Intelligent Robots and Systems (IROS),}  (IEEE,
  2025), pp. 2479--2485.

\bibitem{jahedi2024ccmr}
A.~Jahedi, M.~Luz, M.~Rivinius, and A.~Bruhn, \enquote{Ccmr: High resolution
  optical flow estimation via coarse-to-fine context-guided motion reasoning,}
  in \emph{Proceedings of the IEEE/CVF Winter Conference on Applications of
  Computer Vision,}  (2024), pp. 6899--6908.

\bibitem{sui2022craft}
X.~Sui, S.~Li, X.~Geng, \emph{et~al.}, \enquote{Craft: Cross-attentional flow
  transformer for robust optical flow,} in \emph{Proceedings of the IEEE/CVF
  conference on Computer Vision and Pattern Recognition,}  (2022), pp.
  17602--17611.

\bibitem{dong2024memflow}
Q.~Dong and Y.~Fu, \enquote{Memflow: Optical flow estimation and prediction
  with memory,} in \emph{Proceedings of the IEEE/CVF Conference on Computer
  Vision and Pattern Recognition,}  (2024), pp. 19068--19078.

\bibitem{sun2022skflow}
S.~Sun, Y.~Chen, Y.~Zhu, \emph{et~al.}, \enquote{Skflow: Learning optical flow
  with super kernels,} {\protect\JournalTitle{Advances in Neural Information
  Processing Systems}} \textbf{35}, 11313--11326 (2022).

\bibitem{morimitsu2024recurrent}
H.~Morimitsu, X.~Zhu, X.~Ji, and X.-C. Yin, \enquote{Recurrent partial kernel
  network for efficient optical flow estimation,} in \emph{Proceedings of the
  AAAI Conference on Artificial Intelligence,}  vol.~38 (2024), pp. 4278--4286.

\bibitem{morimitsu2024rapidflow}
H.~Morimitsu, X.~Zhu, R.~M. Cesar, \emph{et~al.}, \enquote{Rapidflow: Recurrent
  adaptable pyramids with iterative decoding for efficient optical flow
  estimation,} in \emph{2024 IEEE International Conference on Robotics and
  Automation (ICRA),}  (IEEE, 2024), pp. 2946--2952.

\bibitem{shi2023videoflow}
X.~Shi, Z.~Huang, W.~Bian, \emph{et~al.}, \enquote{Videoflow: Exploiting
  temporal cues for multi-frame optical flow estimation,} in \emph{Proceedings
  of the IEEE/CVF International Conference on Computer Vision,}  (2023), pp.
  12469--12480.

\bibitem{sun2018pwc}
D.~Sun, X.~Yang, M.-Y. Liu, and J.~Kautz, \enquote{Pwc-net: Cnns for optical
  flow using pyramid, warping, and cost volume,} in \emph{Proceedings of the
  IEEE conference on computer vision and pattern recognition,}  (2018), pp.
  8934--8943.

\bibitem{hui2018liteflownet}
T.-W. Hui, X.~Tang, and C.~C. Loy, \enquote{Liteflownet: A lightweight
  convolutional neural network for optical flow estimation,} in
  \emph{Proceedings of the IEEE conference on computer vision and pattern
  recognition,}  (2018), pp. 8981--8989.

\bibitem{zhao2020maskflownet}
S.~Zhao, Y.~Sheng, Y.~Dong, \emph{et~al.}, \enquote{Maskflownet: Asymmetric
  feature matching with learnable occlusion mask,} in \emph{Proceedings of the
  IEEE/CVF conference on computer vision and pattern recognition,}  (2020), pp.
  6278--6287.

\bibitem{dosovitskiy2015flownet}
A.~Dosovitskiy, P.~Fischer, E.~Ilg, \emph{et~al.}, \enquote{Flownet: Learning
  optical flow with convolutional networks,} in \emph{Proceedings of the IEEE
  international conference on computer vision,}  (2015), pp. 2758--2766.

\bibitem{ilg2017flownet}
E.~Ilg, N.~Mayer, T.~Saikia, \emph{et~al.}, \enquote{Flownet 2.0: Evolution of
  optical flow estimation with deep networks,} in \emph{Proceedings of the IEEE
  conference on computer vision and pattern recognition,}  (2017), pp.
  2462--2470.

\bibitem{hur2019iterative}
J.~Hur and S.~Roth, \enquote{Iterative residual refinement for joint optical
  flow and occlusion estimation,} in \emph{Proceedings of the IEEE/CVF
  conference on computer vision and pattern recognition,}  (2019), pp.
  5754--5763.

\bibitem{zhang2021separable}
F.~Zhang, O.~J. Woodford, V.~A. Prisacariu, and P.~H. Torr, \enquote{Separable
  flow: Learning motion cost volumes for optical flow estimation,} in
  \emph{Proceedings of the IEEE/CVF international conference on computer
  vision,}  (2021), pp. 10807--10817.

\bibitem{godet2021starflow}
P.~Godet, A.~Boulch, A.~Plyer, and G.~Le~Besnerais, \enquote{Starflow: A
  spatiotemporal recurrent cell for lightweight multi-frame optical flow
  estimation,} in \emph{2020 25th International conference on pattern
  recognition (ICPR),}  (IEEE, 2021), pp. 2462--2469.

\bibitem{zheng2022dip}
Z.~Zheng, N.~Nie, Z.~Ling, \emph{et~al.}, \enquote{Dip: Deep inverse patchmatch
  for high-resolution optical flow,} in \emph{Proceedings of the IEEE/CVF
  Conference on Computer Vision and Pattern Recognition,}  (2022), pp.
  8925--8934.

\bibitem{dong2023rethinking}
Q.~Dong, C.~Cao, and Y.~Fu, \enquote{Rethinking optical flow from geometric
  matching consistent perspective,} in \emph{Proceedings of the IEEE/CVF
  Conference on computer vision and pattern recognition,}  (2023), pp.
  1337--1347.

\bibitem{bar2020scopeflow}
A.~Bar-Haim and L.~Wolf, \enquote{Scopeflow: Dynamic scene scoping for optical
  flow,} in \emph{Proceedings of the IEEE/CVF Conference on Computer Vision and
  Pattern Recognition,}  (2020), pp. 7998--8007.

\bibitem{sun2024streamflow}
S.~Sun, J.~Liu, H.~Li, \emph{et~al.}, \enquote{Streamflow: streamlined
  multi-frame optical flow estimation for video sequences,}
  {\protect\JournalTitle{Advances in neural information processing systems}}
  \textbf{37}, 9205--9228 (2024).

\bibitem{morimitsu2025dpflow}
H.~Morimitsu, X.~Zhu, R.~M. Cesar, \emph{et~al.}, \enquote{Dpflow: Adaptive
  optical flow estimation with a dual-pyramid framework,} in \emph{Proceedings
  of the Computer Vision and Pattern Recognition Conference,}  (2025), pp.
  17810--17820.

\bibitem{xu2021high}
H.~Xu, J.~Yang, J.~Cai, \emph{et~al.}, \enquote{High-resolution optical flow
  from 1d attention and correlation,} in \emph{Proceedings of the IEEE/CVF
  International Conference on Computer Vision,}  (2021), pp. 10498--10507.

\bibitem{kong2021fastflownet}
L.~Kong, C.~Shen, and J.~Yang, \enquote{Fastflownet: A lightweight network for
  fast optical flow estimation,} in \emph{2021 IEEE International Conference on
  Robotics and Automation (ICRA),}  (IEEE, 2021), pp. 10310--10316.

\bibitem{shi2022csflow}
H.~Shi, Y.~Zhou, K.~Yang, \emph{et~al.}, \enquote{Csflow: Learning optical flow
  via cross strip correlation for autonomous driving,} in \emph{2022 IEEE
  intelligent vehicles symposium (IV),}  (IEEE, 2022), pp. 1851--1858.

\bibitem{xu2023lla}
J.~Xu, Z.~Lu, and Q.~Liao, \enquote{Lla-flow: A lightweight local aggregation
  on cost volume for optical flow estimation,} in \emph{2023 IEEE International
  Conference on Image Processing (ICIP),}  (IEEE, 2023), pp. 3220--3224.

\bibitem{zhao2022global}
S.~Zhao, L.~Zhao, Z.~Zhang, \emph{et~al.}, \enquote{Global matching with
  overlapping attention for optical flow estimation,} in \emph{Proceedings of
  the IEEE/CVF Conference on Computer Vision and Pattern Recognition,}  (2022),
  pp. 17592--17601.

\bibitem{wang2020displacement}
J.~Wang, Y.~Zhong, Y.~Dai, \emph{et~al.}, \enquote{Displacement-invariant
  matching cost learning for accurate optical flow estimation,}
  {\protect\JournalTitle{Advances in Neural Information Processing Systems}}
  \textbf{33}, 15220--15231 (2020).

\end{thebibliography}
\end{document}